\documentclass[10pt,twocolumn,letterpaper]{article}

\usepackage[pagenumbers]{iccv} 

\usepackage{graphicx}
\usepackage{booktabs}
\usepackage{multirow}
\usepackage{longtable}
\usepackage{comment}
\usepackage{placeins}
\usepackage{graphicx}
\usepackage{titletoc}
\usepackage{tabularx}
\usepackage{amsmath}
\usepackage{float}
\usepackage[accsupp]{axessibility}  

\usepackage{adjustbox}

\definecolor{iccvblue}{rgb}{0.21,0.49,0.74}
\usepackage[pagebackref,breaklinks,colorlinks,allcolors=iccvblue]{hyperref}

\def\paperID{8610} 
\def\confName{ICCV}
\def\confYear{2025}

\newcommand{\YSR}[1]{\textcolor{blue}{#1}}

\title{
Punching Bag vs. Punching Person: Motion Transferability in Videos
}
\author{Raiyaan Abdullah\textsuperscript{1}\\
{\tt\small raiyaanabdullah@gmail.com}
\and
Jared Claypoole\textsuperscript{2}\\
{\tt\small jared.claypoole@sri.com}
\and
Michael Cogswell\textsuperscript{2}\\
{\tt\small michael.cogswell@sri.com}
\and
Ajay Divakaran\textsuperscript{2}\\
{\tt\small ajay.divakaran@sri.com}
\and
Yogesh Rawat\textsuperscript{1}\\
{\tt\small yogesh@crcv.ucf.edu}
\and
\textsuperscript{1}\small Center for Research in Computer Vision, University of Central Florida
\textsuperscript{2}\small Center for Vision Technology, SRI International\\
\normalsize \textbf{\href{http://raiyaan-abdullah.github.io/Motion-Transfer-webpage/}{Project page}}
}

\begin{document}
\twocolumn[{
\maketitle
\begin{center}
    \vspace{-20pt}
    \centering 
    \captionsetup{type=figure}
    \includegraphics[width=1\textwidth]{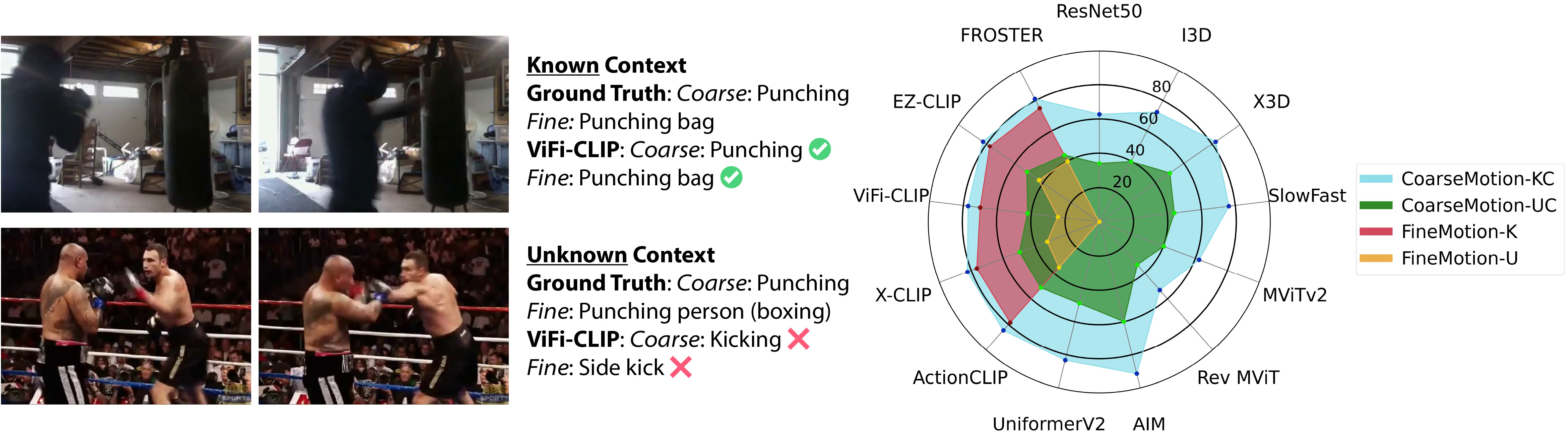}
    \vspace{-15pt}
    \caption{\textbf{Overview of motion transferability:} \textit{(left)} Models fail to transfer high-level coarse motion understanding (`Punching') to unknown context not seen during training. State-of-the-art multimodal models like ViFi-CLIP fail to understand unknown fine motions such as (`Punching person - boxing') as well. \textit{(right)} Average detection accuracy (across three datasets) illustrating the performance gap between CoarseMotion-KnownContext and CoarseMotion-UnknownContext from blue to green, as well as FineMotion-Known and FineMotion-Unknown from red to yellow. This highlights that all models have limitations in transferring motion concepts to novel scenarios.
    }
\label{fig:teaser}
\end{center}
}]
\begin{abstract}

Action recognition models demonstrate strong generalization, but can they effectively transfer high-level motion concepts across diverse contexts, even within similar distributions? For example, can a model recognize the broad action \textit{``punching''} when presented with an unseen variation such as \textit{``punching person''}? To explore this, we introduce a motion transferability framework with three datasets: (1) \textbf{Syn-TA}, a synthetic dataset with 3D object motions; (2) \textbf{Kinetics400-TA}; and (3) \textbf{Something-Something-v2-TA}, both adapted from natural video datasets. We evaluate 13 state-of-the-art models on these benchmarks and observe a significant drop in performance when recognizing high-level actions in novel contexts. Our analysis reveals: 1) Multimodal models struggle more with fine-grained unknown actions than with coarse ones; 2) The bias-free \textbf{Syn-TA} proves as challenging as real-world datasets, with models showing greater performance drops in controlled settings; 3) Larger models improve transferability when spatial cues dominate but struggle with intensive temporal reasoning, while reliance on object and background cues hinders generalization. We further explore how disentangling coarse and fine motions can improve recognition in temporally challenging datasets. We believe this study establishes a crucial benchmark for assessing motion transferability in action recognition.  Datasets and relevant code: \href{https://github.com/raiyaan-abdullah/Motion-Transfer}{https://github.com/raiyaan-abdullah/Motion-Transfer}.

\vspace{-10pt}
\end{abstract}   
\section{Introduction}
\label{sec:intro}
Activity recognition is a key area of video understanding, focusing on identifying motions within videos by extracting meaningful insights from complex data. This challenging task involves handling temporal redundancy \cite{vlad_temporal_redundancy}, long-term dependencies \cite{rnn_long_term_dependency}, varying viewpoints \cite{pyramid_transformer_viewpoint_scale}, diverse subjects/objects \cite{object_action_relation}, and cluttered backgrounds \cite{crowd_clutter_action4d}. Recent unimodal and multimodal activity recognition models have achieved strong performance in supervised and few-shot settings, with multimodal models ~\cite{actionclip,xclip,vificlip,ezclip, froster} also excelling in zero-shot generalization.

Despite progress in action recognition, questions remain about models' ability to understand high-level motions and generalize them to novel contexts. Existing studies on zero-shot learning, base-to-novel transfer, and domain adaptation reveal challenges with distribution shifts and variations in view, environment, outcome, objects, or people, and propose mitigation methods. Yet it remains unclear whether a model trained on ``\textit{punching bag}'' videos can recognize ``\textit{punching}'' in contexts such as ``\textit{punching person}'' or identify coarse motions while disregarding extra contextual details, a skill that humans can learn \cite{learningthattransfers}. As illustrated in \cref{fig:teaser} (left), models often misclassify actions within similar distributions due to context-dependent biases. Unlike existing evaluation protocols that test generalization across different distributions, our study focuses on whether models can overcome fine-class bias (e.g. recognizing ``punching'' beyond ``punching bag'') rather than coarse-class bias or scenario bias (e.g. temporal redundancy, long-term dependencies, diverse subjects, cluttered backgrounds). Fine-class persists because datasets share similar conditions, hindering true motion generalization.

We present a systematic study of this form of activity transferability by proposing three benchmark datasets. Firstly, to investigate transferability in a controlled setting, we design ``Syn-TA'' consisting of videos rendered in Blender~\cite{blender}, featuring 3D objects performing standard motions. Next, we propose two real-world benchmarks, ``Kinetics400 - Transferable Activity'' (K400-TA) and ``Something-something-v2 - Transferable Activity'' (SSv2-TA), where we utilize two well-known large-scale datasets~\cite{kinetics, somethingsomething}. Each dataset is structured with a hierarchy of high-level ``coarse'' classes associated with groups of lower-level ``fine'' motion classes constituting different contexts. We split each dataset into two sets with same coarse actions but disjoint fine-grained actions. Models are trained separately on each set and evaluated on known and unknown splits to measure their generalization ability of high-level motions in new scenarios. We evaluate 13 state-of-the-art unimodal and multimodal models on these datasets to systematically assess their ability to transfer learned motions across new contexts. Multimodal models match video embeddings to text descriptions, enabling them to classify unseen fine-grained motions without retraining, a capability unimodal models lack. Thus, we evaluate unimodal models on coarse motions only, and multimodal models on both coarse and fine motions.

Our experiments confirm that existing models struggle to generalize high-level motions across contexts, showing a significant drop when transitioning from known to unknown scenarios (\cref{fig:teaser}, right). 
Multimodal models further decline on fine-context activities due to finer distinctions and a larger number of classes. Our study reveals several key findings: 1) the synthetic dataset with a controlled setting proves more challenging than K400-TA, 2) compared to real-world datasets, it also exhibits a higher drop from coarse to fine motions in unknown scenarios,  
3) larger models improve generalization for spatially inferred motions but not for those requiring deeper temporal understanding,
and 4) leveraging controlled experiments, we demonstrate through ``Syn-TA'' that background texture and information influence model performance in capturing motion accurately, underscoring the role of context. In view of these findings, 
we explore disentangling fine-context motions during training to enhance high-level understanding and transferability.


Our main contributions are as follows:
\begin{itemize}
\item We formalize studying motion transferability by introducing an intra-dataset coarse-to-fine hierarchy that is distinct from existing setups. Based on this, we propose three benchmark datasets: ``Syn-TA'', ``K400-TA'', and ``SSv2-TA'', encompassing both controlled bias-free and real-world biased settings.
\item We systematically evaluate 13 unimodal and multimodal models, showing their performance always deteriorates between known and unknown contexts and, in most cases, between coarse and fine context classes. We provide further insights into the challenges of a bias-free setting, the influence of model architecture, and the performance drop caused by background complexity.
\item We propose a disentanglement strategy that utilizes information from fine-context features during training. This improves transferability of high-level coarse actions across contexts in temporal datasets.
\end{itemize}

\section{Related Work}
\label{sec:relwork}
\paragraph{Video Action Recognition:}
Action recognition methods detect motion in videos using spatial-temporal cues, including 3D CNN based models~\cite{tran2015learning}, capsule networks \cite{capsulenet} and vision transformers~\cite{vit}.~\cite{uniformer} combines 3D convolutions with self-attention. ~\cite{mvit, mvitv2} show that multi-scale features effectively capture video signals, while ~\cite{aim} employs adapters, freezing pre-trained image model parameters. However, these unimodal approaches classify only motions seen during training.
Zero-shot methods~\cite{xu2015semantic,lin2022crossmodalrepresentationlearningzeroshot,brattoli2020rethinking,rethinkingzeroshot,reformulatingzeroshot} match flexible text embeddings with visual features in a high-dimensional space, exemplified by CLIP~\cite{clip}. Richer text descriptions~\cite{gowda2024tellingstoriescommonsense} enable models to facilitate stronger video-text alignment \cite{clip, eilev, hierarq, videochat}. ~\cite{towards_fair_evaluation,gowda2021newsplitevaluatingtrue,chen2021elaborativerehearsalzeroshotaction} have explored various setups of zero-shot action recognition.
Several works have adapted CLIP to video tasks such as retrieval, video question answering, and activity recognition \cite{videoclip, actionclip, xclip, vificlip, ezclip, t2l, froster}. These methods utilize prompting, fine-tuning, adapter modules, and distillation to match video embeddings with textual action descriptions and achieve strong few shot, zero shot, and base-to-novel transfer on standard datasets.

\noindent
\textbf{Deeper Understanding of Models:}  
Existing works pursue deeper model understanding by tackling issues such as robustness, low-resolution, noisy labels, multi-label settings, and distribution shift~\cite{largescalerobustness,tinyvirat,noisyactions,ucfmama}. They also investigate how scene and object cues influence understanding and steer motion prediction ~\cite{schiappa2024probing,crepe,azad2025understanding,object_action_google}. To capture compositional and hierarchical structure, models form representations that learn subject–object relations and organize actions from coarse to fine \cite{somethingelse,hierarchical,verbs_in_action}. Disentanglement methods separate interacting factors such as human–object interaction streams or view-specific cues for more precise reasoning \cite{human_object_disentangle,dvanet}. These studies typically assess performance using either standard action recognition metrics or object detection metrics.

\begin{figure}[t!]
\centering
\includegraphics[width=.9\linewidth]{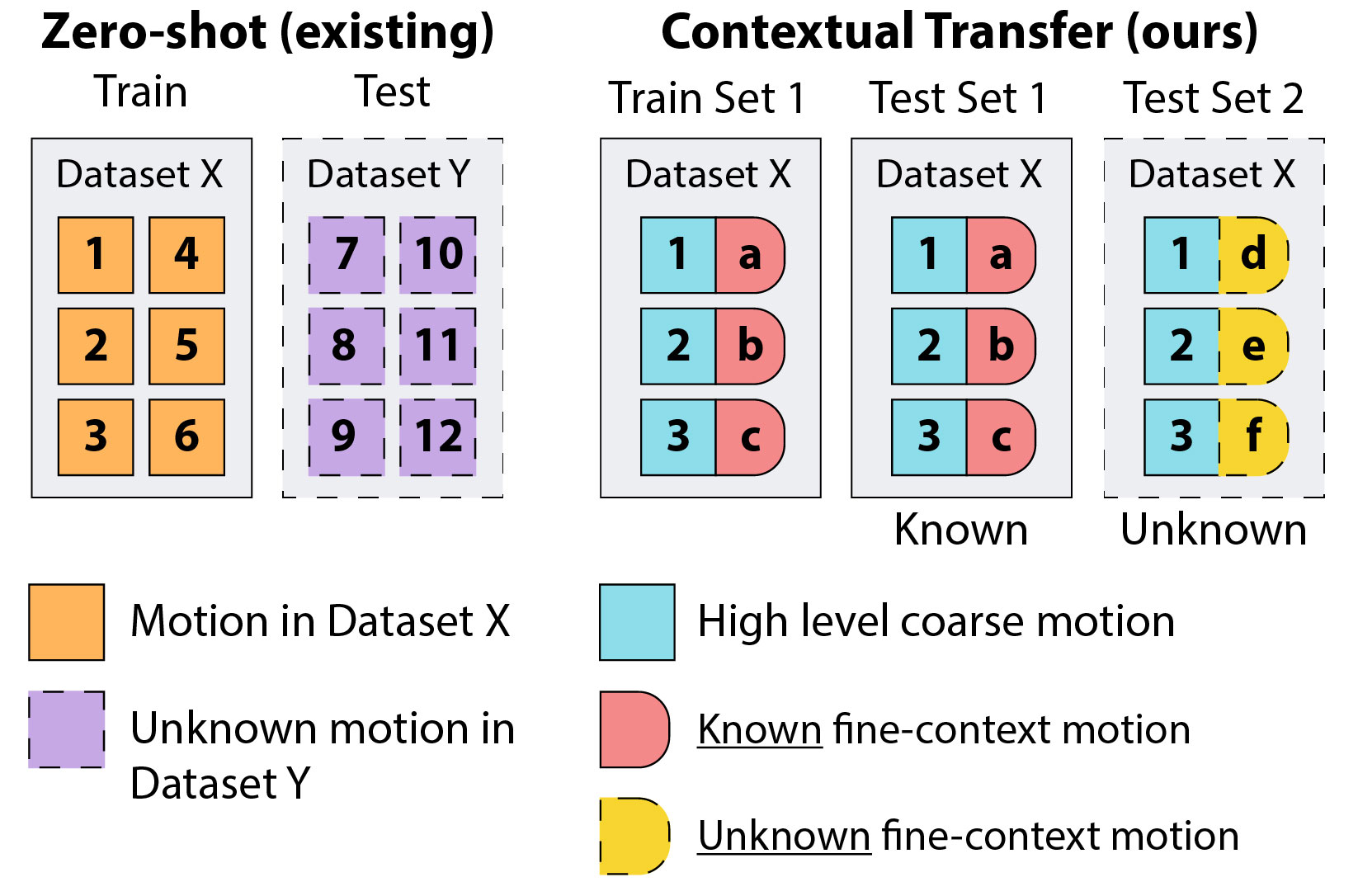}
\caption{
\textbf{\textit{Comparison with zero-shot setup:}}
We categorize motions into high-level (coarse) groups, creating two sets with similar coarse motions but varied fine details. After training on Train Set 1 (or Set 2), we evaluate performance on known motions in Test Set 1 (or Set 2) and unseen variations in Test Set 2 (or Set 1).
}
 \label{fig:zero_vs_ours}
 \vspace{-15pt}
\end{figure}

\noindent
\textbf{Domain Adaptation and Generalization:}
Base-to-novel and zero-shot settings focus on unseen classes and thus do not measure transfer of motion knowledge across similar classes. Domain adaptation and domain generalization address distribution shifts between similar classes: domain adaptation uses unlabeled target videos during training \cite{temporalattentive, interactbeforealign, stcontrastive, learningcrossmodal, relativenormalignment} while domain generalization has no target data. Prior works evaluate motion generalization across datasets such as UCF to HMDB, and K400 to Drone videos or within datasets such as NTU across viewpoints, SSv2 across different outcomes, EpicKitchens across kitchens, and ARGO1M across scenarios \cite{videodg,epickitchensuda,stdn,cookitalyindia,necdrone}. \cite{vlmodelactioneffectiveness} compares VL models on coarse actions from Penn Action \cite{pennaction} and fine-grained actions from Smarthome-CS \cite{toyotasmarthome} but lacks a clear intra-dataset hierarchy.

\noindent Our approach introduces a novel evaluation framework by structuring datasets into a coarse-to-fine hierarchy to assess model recognition of coarse actions across varied fine contexts. Unlike zero-shot action recognition and cross-dataset domain generalization, which involve novel classes or different datasets, we evaluate generalization within the same dataset. Base-to-novel generalization splits classes by frequency without semantic organization, and intra-dataset domain generalization introduces scenario biases. Instead, we systematically group fine-context motions into broader categories, focusing on \textit{which context} defines the action rather than \textit{location, scenario, or distribution shift}. To the best of our knowledge, while such fine-level distinctions may be present in some works, this critical aspect of generalization has not been systematically explored. As illustrated in \cref{fig:zero_vs_ours}, our approach focuses on evaluation of fine-context biases, isolating core action understanding from extraneous factors. While existing methods target distribution shifts across datasets, we emphasize on motion generalization within the same or across different distributions.
\vspace{-5pt}

\section{Benchmark Details and Disentanglement}
\label{sec:benchmark_details}
\begin{table}
\scriptsize
\centering
\begin{adjustbox}{width=\linewidth,keepaspectratio}
\begin{tabular}{l|c|c|c}
\toprule
\textbf{Dataset} & \textbf{Syn-TA} & \textbf{K400-TA} & \textbf{SSv2-TA} \\ \hline
 & \textbf{Set 1/Set 2} & \textbf{Set 1/Set 2} & \textbf{Set 1/Set 2} \\ 
\hline
{\# Coarse classes} & 20 & 41 & 26 \\              
{\# Fine classes} & 53/47 & 111/94 & 81/68 \\                      
{\# Train videos} & 3180/2820 & 73312/56291 & 78229/63611 \\                      
{\# Test videos} & 2120/1880 & 5616/4664 & 11134/9241 \\      
{\# Total videos} & 10000 & 139883 & 162215 \\  
\bottomrule
\end{tabular}
\end{adjustbox}
\vspace{-5pt}
\caption{\textbf{\textit{Details of datasets:}} Overview of the number of classes and videos in Syn-TA, K400-TA, and SSv2-TA. 
}
\vspace{-5pt}
\label{tab:dataset_overview}
\end{table}

\begin{figure}[t!]
\centering
\vspace{-10pt}
  \includegraphics[width=1\linewidth]{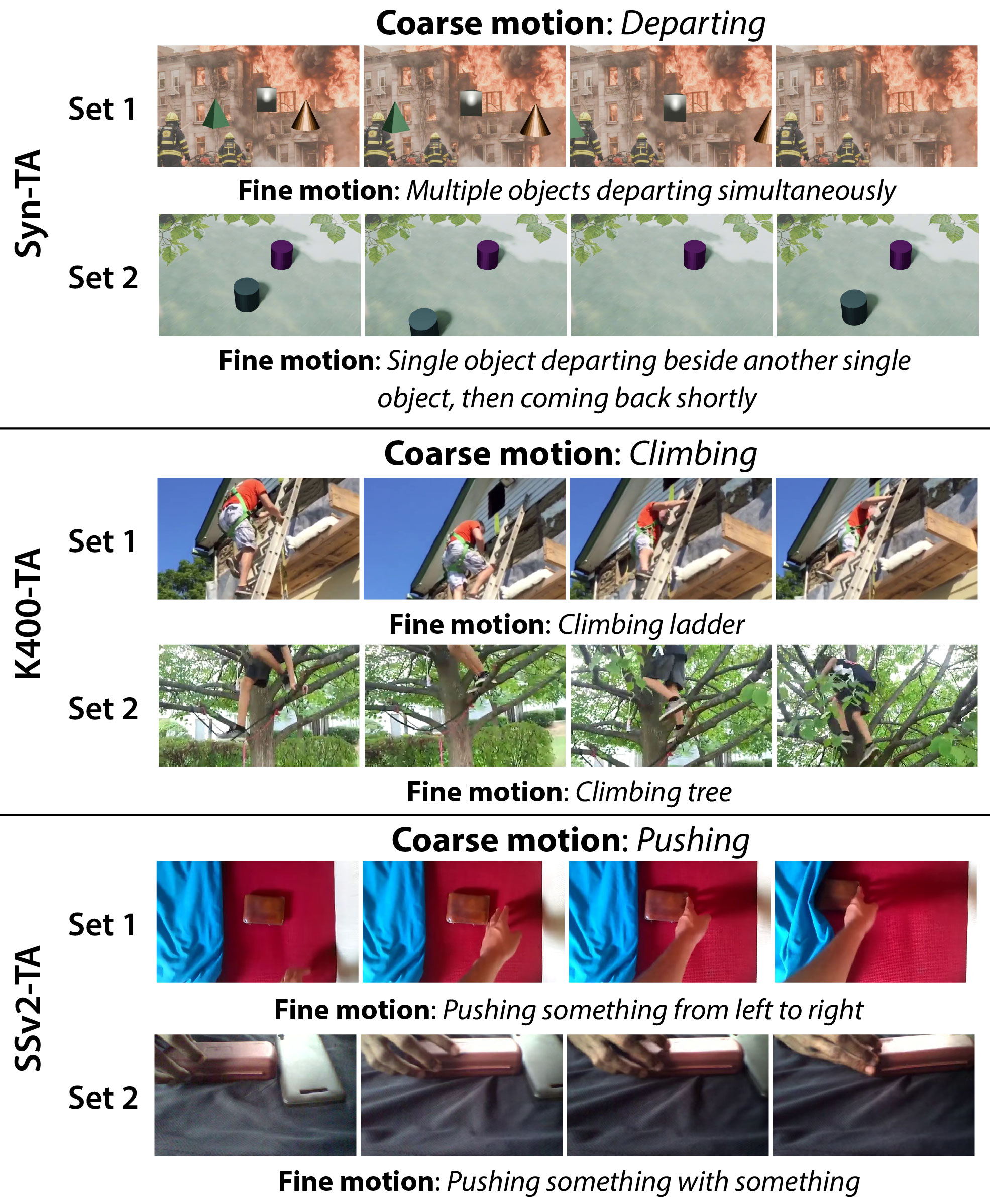}
\caption{\textbf{\textit{Preview of benchmark datasets:}} Each dataset is divided into Set 1 and Set 2, where both share similar coarse motions but differ in fine motions. Examples of coarse and fine classes from Syn-TA, K400-TA, and SSv2-TA are provided.
}
\label{fig:all_dataset_example}
\vspace{-20pt}
\end{figure}

To investigate the transferability of motion into unknown contexts, we propose three different benchmark datasets: Syn-TA, K400-TA, and SSv2-TA. In the following, we discuss details about curating these datasets and describe our disentanglement strategy for improving transferability.
\subsection{Dataset Construction}
Existing large-scale action-recognition datasets treat all motions at a uniform level of detail. Our proposed datasets introduce both high-level coarse classes and lower-level fine classes similar to \cite{hierarchical}. The coarse classes, such as ``\textit{Falling}'' and ``\textit{Pushing}'', represent broad motions, capturing the general nature without specific contextual information. In contrast, fine classes such as ``\textit{Single object falling at an angle}'' or ``\textit{Pushing something from right to left}'' specify both the high-level motion and detailed context about how the motion is performed. This context may correspond to different objects but often does not. As described in \cref{fig:zero_vs_ours}, alternative approaches do not capture the generalization of known classes across mutually exclusive contexts.

Given a large video action dataset $D$ with $N$ classes, we adapt it to our problem by identifying a subset $S \subset N$ of fine classes and grouping them into $C$ coarse classes performing similar motions. We split the fine classes in $S$ into sets $S1$ and $S2$, ensuring all high-level coarse classes are represented in both sets and each set contains approximately a similar number of fine classes for each coarse class. The two sets contain a similar number of coarse classes but different number of fine classes. The corresponding videos in the training set $D_{Train}$ and validation set $D_{Test}$ of $S$ are then split into two groups ($D_{Train_{S1}}$, $D_{Train_{S2}}$, and $D_{Test_{S1}}$, $D_{Test_{S2}}$) accordingly. We train the models with videos in $D_{Train_{S1}}$ and evaluate both known $D_{Test_{S1}}$ and unknown context $D_{Test_{S2}}$ and vice versa.

\subsection{Proposed Datasets}
To explore high-level motions in known and unknown contexts in a controlled setting, we introduce a new synthetic dataset: ``Syn-TA''. We also adapt two existing large-scale datasets containing real-world videos: ``SSv2-TA'' and ``K400-TA''. The latter captures diverse real-world scenarios with biases such as overlapping objects, imbalanced fine motions per coarse class, and uneven sample distributions. Examples can be found in \cref{fig:all_dataset_example}, and statistics are reported in \cref{tab:dataset_overview}, with more details in supplementary.
\vspace{-15pt}
\paragraph{Syn-TA:}
We introduce ``Syn-TA'' to investigate motion generalization in a synthetic controlled environment. This enables us to precisely maintain motion consistency while varying actors and contexts, which is not possible with real-world datasets. This allows a clearer evaluation of a model's ability to generalize high-level motion concepts. Using Blender~\cite{blender}, we generate videos of 3D objects such as cubes, spheres, cylinders, etc., performing various motions like \textit{``appearing by increasing opacity''} or \textit{``arriving among many objects''} against realistic background images. 

\noindent ``Syn-TA'' comprises 20 coarse classes and 100 fine classes. The coarse activities are simple motions that can be animated through these 3D objects. The fine motions are characterized by more specific details, such as precise path, object positioning, and the number of objects involved. Each coarse class is associated with 4 to 8 fine motions, with each fine motion having 60 videos in $D_{Train}$ and 40 in $D_{Test}$. Object shapes, colors, and camera positions are varied randomly in each video to create variations. Each of the two sets has unique background images and object shapes to prevent the model from exploiting these cues. This balanced low-bias design enables clearer evaluation of a model’s ability to transfer high-level motion concepts.
\vspace{-12pt}
\paragraph{Kinetics400 - Transferable Activity (K400-TA):} 
Our first realistic dataset is adapted from Kinetics400~\cite{kinetics}, where we use 205 fine classes from the original 400 classes in the dataset and group them into 41 coarse classes, omitting those too general for falling under a clearly defined coarse class. Recognizing motions in this dataset relies heavily on spatial cues. Objects, actors, and scene details often define the class, with fine-grained context primarily deriving from them. For example, under the coarse class \textit{``Cleaning''}, fine classes include \textit{``Cleaning floor''} and \textit{``Cleaning gutters''}. Some fine classes were renamed from their original label to better capture context.
\vspace{-12pt}
\paragraph{Something-something-v2 - Transferable Activity (SSv2-TA):}
 In Something-something-v2~\cite{somethingsomething} the descriptions of the classes are structured as ``coarse activity + fine context''. For example, the class \textit{``Pushing something from left to right''} contains both the coarse activity \textit{``Pushing''} and fine context \textit{``something from left to right''}, so coarse and fine classes naturally emerge from the data. We selected 26 coarse classes that correspond to multiple fine classes, utilizing 149 of the original 174 motion classes as the fine classes. Understanding temporal dynamics is crucial for models to correctly recognize motions in this dataset.

\begin{table*}[!ht]
\small
\centering
\resizebox{17.5cm}{!}{%
\begin{tabular}{l|cccc|cccc|cccc}
\toprule
\textbf{Model} 
  & \multicolumn{4}{c|}{\textbf{Syn-TA}} 
  & \multicolumn{4}{c|}{\textbf{K400-TA}} 
  & \multicolumn{4}{c}{\textbf{SSv2-TA}} \\
\hline
  & \textbf{Known \textuparrow} & \textbf{Unknown \textuparrow} & \textbf{D\textsubscript{abs} \textdownarrow} & \textbf{HM \textuparrow} 
  & \textbf{Known \textuparrow} & \textbf{Unknown \textuparrow} & \textbf{D\textsubscript{abs} \textdownarrow} & \textbf{HM \textuparrow} 
  & \textbf{Known \textuparrow} & \textbf{Unknown \textuparrow} & \textbf{D\textsubscript{abs} \textdownarrow} & \textbf{HM \textuparrow} \\
\hline
\multicolumn{13}{c}{\textbf{Unimodal Models}} \\
\hline
ResNet50~\cite{resnet50}      & 66.66 & 29.93 & 36.73 & 41.30 & 76.49 & 46.21 & 30.28 & 57.59 & 45.07 & 26.08 & 18.99 & 33.01 \\
I3D~\cite{i3d}                & 80.50 & 37.51 & 42.99 & 51.17 & 76.89 & 47.25 & 29.63 & 58.49 & 59.60 & 34.40 & 25.20 & 43.53 \\
X3D~\cite{x3d}                & 93.71 & 58.45 & 35.25 & 71.79 & 81.23 & 49.88 & 31.35 & 61.78 & 72.73 & 41.81 & 30.92 & 53.05 \\
SlowFast~\cite{slowfast}      & 89.27 & 46.86 & 42.41 & 61.45 & 81.70 & 50.33 & 31.37 & 62.26 & 57.67 & 35.15 & 22.51 & 43.60 \\
MViTv2~\cite{mvitv2}          & 63.69 & 43.23 & \textbf{\underline{20.46}} & 51.50 & 68.88 & 45.06 & 23.81 & 54.47 & 54.31 & 32.37 & 21.93 & 40.49 \\
Rev-MViT~\cite{rev_mvit}       & 65.53 & 38.02 & 27.51 & 47.98 & 59.40 & 40.54 & \textbf{\underline{18.86}} & 48.16 & 34.64 & 21.72 & \textbf{\underline{12.92}} & 26.68 \\
AIM~\cite{aim}                & \textbf{\underline{99.13}} & \textbf{\underline{70.16}} & 28.97 & \textbf{\underline{82.17}} & 95.04 & 63.73 & 31.31 & 76.29 & \textbf{\underline{79.94}} & \textbf{\underline{45.82}} & 34.12 & \textbf{\underline{58.18}} \\
UniformerV2~\cite{uniformerv2}& 97.96 & 51.20 & 46.76 & 67.25 & 93.56 & 62.29 & 31.27 & 74.77 & 58.16 & 33.20 & 24.96 & 42.25 \\
\hline
\multicolumn{13}{c}{\textbf{Multimodal Models}} \\
\hline
ActionCLIP~\cite{actionclip}  & 96.29 & 55.33 & 40.95 & 70.27 & 93.24 & 62.24 & 31.00 & 74.60 & 64.10 & 36.66 & 27.44 & 46.56 \\
X-CLIP~\cite{xclip}           & 85.04 & 47.83 & 37.21 & 61.22 & 92.69 & 61.47 & 31.22 & 73.90 & 69.49 & 40.10 & 29.39 & 50.74 \\
ViFi-CLIP~\cite{vificlip}      & 79.67 & 35.46 & 44.21 & 49.01 & 93.24 & 60.44 & 32.80 & 73.31 & 58.69 & 30.69 & 27.99 & 40.22 \\
EZ-CLIP~\cite{ezclip}          & 98.30 & 52.43 & 45.87 & 68.38 & 86.88 & 66.70 & 20.18 & 75.43 & 62.55 & 34.84 & 27.70 & 44.72 \\
FROSTER~\cite{froster}         & 89.42 & 31.80 & 57.61 & 46.91 & \textbf{\underline{95.99}} & \textbf{\underline{69.23}} & 26.76 & \textbf{\underline{80.42}} & 57.65 & 30.68 & 26.97 & 39.98 \\
\hline
\multicolumn{13}{c}{\textbf{Domain Generalization Methods}} \\
\hline
VideoDG~\cite{videodg} & 98.07 & 43.43 & 54.64 & 60.17 & 86.11 & 53.95 & 32.15 & 66.27 & 57.25 & 31.54 & 25.71 & 40.63 \\
STDN~\cite{stdn} & 70.66 & 23.97 & 46.69 & 35.72 & 68.11 & 46.10 & 22.01 & 54.89 & 35.93 & 22.31 & 13.62 & 27.51 \\
CIR~\cite{cookitalyindia} & 60.13 & 9.59 & 50.54 & 16.41 & 68.53 & 12.66 & 55.87 & 21.34 & 48.01 & 31.97 & 16.04 & 38.37 \\
\bottomrule
\end{tabular}}
\vspace{-5pt}
\caption{\textbf{\textit{Benchmark for coarse actions:}} Absolute drop and harmonic mean of known (CoarseMotion-KC) and unknown (CoarseMotion-UC) accuracies (average of two sets) for \textbf{coarse activities} across all datasets. 
}
\label{tab:benchmark_results_coarse}
\end{table*}

\subsection{Evaluation setup}

\begin{figure*}[!ht]
\centering
\includegraphics[width=0.9\textwidth]{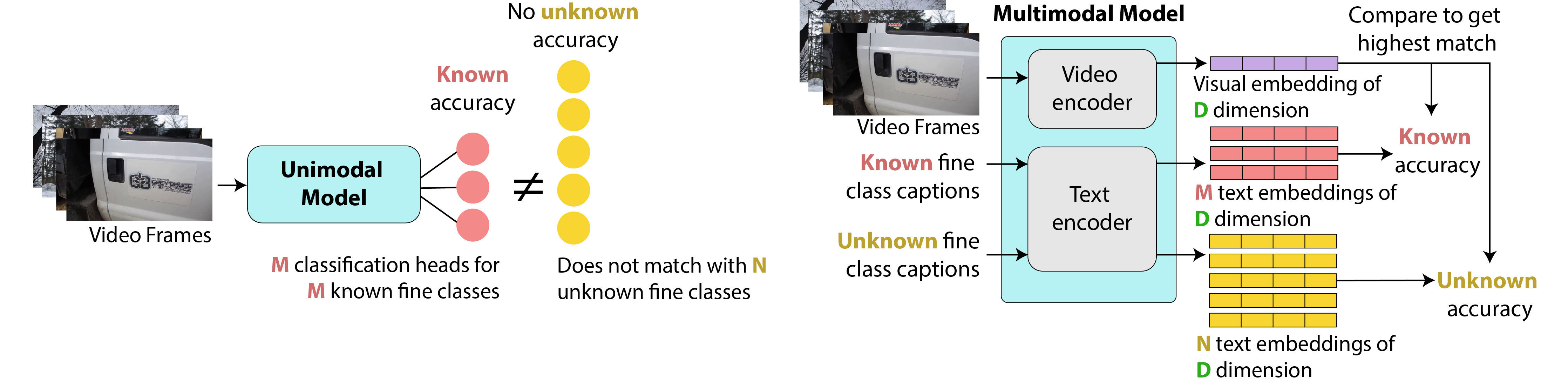}
\vspace{-10pt}
\caption{\textbf{\textit{Unimodal vs. multimodal models:}} Evaluation of fine motions in unimodal models is not possible due to a fixed number of classification heads (left). Multimodal models avoid this limitation since visual and text embeddings both have dimension D (right).
}
\vspace{-15pt}
\label{fig:uni_vs_multi_archi_compare}
\end{figure*}

To assess whether unimodal and multimodal models genuinely understand high-level motions, we design a structured training and evaluation process. As seen in \cref{fig:uni_vs_multi_archi_compare}, unimodal models rely on fixed classification heads for classifying motions. Multimodal models match video embeddings with text embeddings of motion descriptions and can recognize unseen fine motions without retraining, unlike unimodal models. We therefore evaluate unimodal models on coarse motions only and multimodal models on both coarse and fine motions.  

\subsection{Evaluation metrics}
For coarse motions, we train the unimodal or multimodal model on Set 1’s training data $D_{Train_{S1}}$ and evaluate on both $D_{Test_{S1}}$ (CoarseMotion-KnownContext) and $D_{Test_{S2}}$ (CoarseMotion-UnknownContext). To ensure robustness, we repeat this with Set 2, training on $D_{Train_{S2}}$ and evaluating on $D_{Test_{S2}}$ (CoarseMotion-KC) and $D_{Test_{S1}}$ (CoarseMotion-UC). If the model truly understands the high-level concept, its performance on coarse classes should remain consistent across CoarseMotion-KC and CoarseMotion-UC. To quantify generalization across different contexts, we compute:\\
\textbf{Absolute Drop}: The difference in accuracy between known and unknown contexts: $D_{abs} = |Known - Unknown|$\\
\textbf{Harmonic Mean (HM)}: 
The harmonic mean of known and unknown accuracy to fairly balance performance.\\
As multimodal models can also evaluate unknown fine motions, we retrain each model separately on the same videos using fine captions. After training on $D_{Train_{S1}}$, we first evaluate on $D_{Test_{S1}}$ (FineMotion-Known) which contains fine classes seen in training. Then it is evaluated on $D_{Test_{S2}}$ (FineMotion-Unknown), where the number of fine motions in Set 2 differs. The procedure is repeated with $D_{Train_{S2}}$.
\vspace{-5pt}

\begin{table*}[t!]
\small
\centering
\resizebox{17.5cm}{!}{%
\begin{tabular}{l|cccc|cccc|cccc}
\toprule
\textbf{Model} & \multicolumn{4}{c|}{\textbf{Syn-TA}} & \multicolumn{4}{c|}{\textbf{K400-TA}} & \multicolumn{4}{c}{\textbf{SSv2-TA}} \\
\cline{2-13}
  & \textbf{Known \textuparrow} & \textbf{Unknown \textuparrow} & \textbf{D\textsubscript{abs} \textdownarrow} & \textbf{HM \textuparrow} 
  & \textbf{Known \textuparrow} & \textbf{Unknown \textuparrow} & \textbf{D\textsubscript{abs} \textdownarrow} & \textbf{HM \textuparrow} 
  & \textbf{Known \textuparrow} & \textbf{Unknown \textuparrow} & \textbf{D\textsubscript{abs} \textdownarrow} & \textbf{HM \textuparrow} \\
\midrule
ActionCLIP~\cite{actionclip} & 88.01 & \textbf{\underline{38.81}} & \textbf{\underline{49.19}} & \textbf{\underline{53.85}} & 87.75 & 41.52 & 46.23 & 56.20 & 59.72 & 25.84 & 33.88 & 36.03 \\
X-CLIP~\cite{xclip}           & 75.20 & 22.90 & 52.29 & 34.98 & \textbf{\underline{89.06}} & 48.11 & 40.95 & 62.37 & \textbf{\underline{65.31}} & 26.53 & 38.78 & 37.69 \\
ViFi-CLIP~\cite{vificlip}     & 69.27 & 19.91 & 49.36 & 30.79 & 88.91 & 26.70 & 62.21 & 40.97 & 52.13 & 26.28 & 25.85 & 34.93 \\
EZ-CLIP~\cite{ezclip}         & \textbf{\underline{89.54}} & 24.89 & 64.64 & 38.71 & 83.76 & 73.95 &  \textbf{\underline{9.81}} & 78.47 & 59.83 & \textbf{\underline{29.73}} & 30.09 & \textbf{\underline{39.70}} \\
FROSTER~\cite{froster}        & 85.44 & 20.68 & 64.76 & 33.26 & 88.93 & \textbf{\underline{74.11}} & 14.82 & \textbf{\underline{80.81}} & 50.34 & 24.99 & \textbf{\underline{25.35}} & 33.34 \\
\bottomrule
\end{tabular}}
\caption{\textbf{\textit{Benchmark results for fine actions:}} Absolute drop and harmonic mean of known (FineMotion-K) and unknown (FineMotion-U) accuracies (average of two sets) for \textbf{fine motions} across all datasets.}
\label{tab:benchmark_results_fine}
\vspace{-10pt}
\end{table*}

\subsection{Disentanglement of Coarse and Fine}
\label{sec:disentanglement}
\vspace{-5pt}
Our insights in \cref{sec:exp_and_results} reveal that models rely on fine-grained cues in the scene. We investigate whether these distinct features can be learned during training to improve transferability of actions. 
We propose a simple strategy where the model predicts both high-level concepts and fine-grained details. 
This can be achieved by additional layers towards the end of the encoder, creating two distinct branches: one specializing in high-level (coarse) concepts (e.g., \textit{``Pushing''}) and the other focusing on fine-grained contexts (e.g., \textit{``Pushing something from left to right''}). Since the earlier layers capture low and mid-level features, this design allows the final branches to specialize in refining their respective representations. To strengthen coarse reasoning, we integrate fine features into the coarse branch via residual connections after each block, facilitating the incorporation of necessary scene details. We apply this to EZ-CLIP \cite{ezclip}; detailed architecture is in supplementary.
\vspace{-5pt}

\section{Experiments and Results}
\label{sec:exp_and_results}

We experimented with 13 models (8 unimodal and 5 multimodal). For unimodal models we included traditional CNNs such as ResNet50~\cite{resnet50}, I3D~\cite{i3d}, X3D~\cite{x3d}, and SlowFast~\cite{slowfast}. We also used more recent transformer-based unimodal models like MViTv2~\cite{mvitv2}, Rev-MViT~\cite{rev_mvit}, UniformerV2~\cite{uniformerv2}, and AIM~\cite{aim}. Among multimodal models, we experimented with ActionCLIP~\cite{actionclip}, XCLIP~\cite{xclip}, ViFi-CLIP~\cite{vificlip}, EZ-CLIP~\cite{ezclip}, and FROSTER~\cite{froster}. We also investigate how various domain generalization methods perform \cite{videodg,stdn,cookitalyindia} on our datasets. Implementation details can be found in the supplementary.

\begin{figure*}[!ht]
\centering
\includegraphics[width=0.9\textwidth]{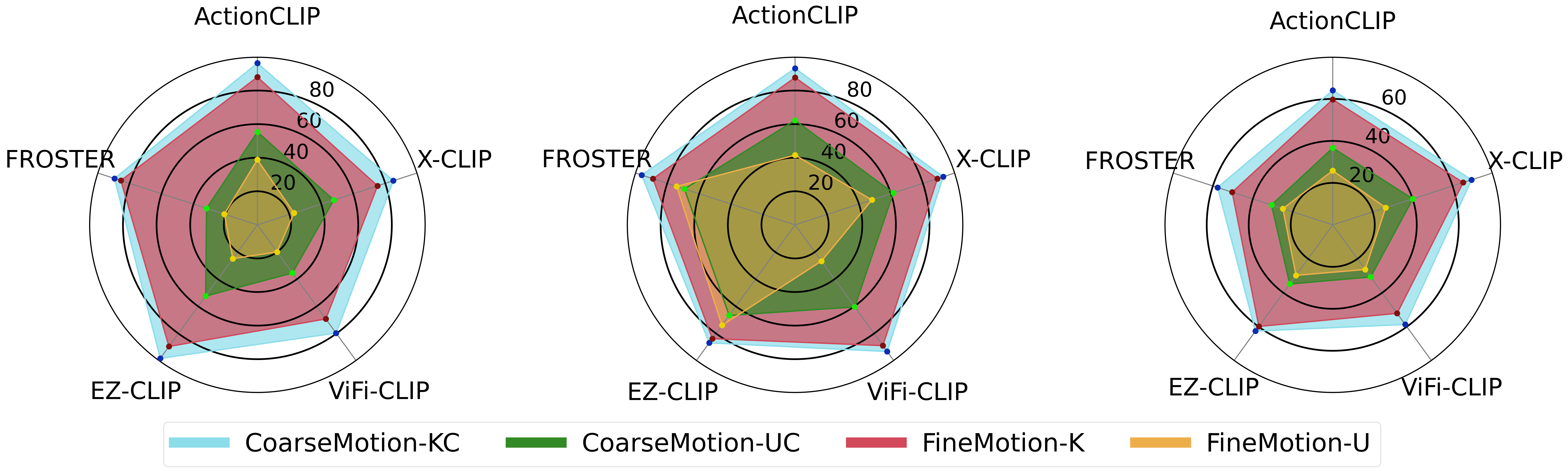}
\vspace{-5pt}
\caption{\textbf{\textit{Comparison for multimodal models:}} Left: Syn-TA, middle: K400-TA, and right: SSv2-TA. Across all datasets, a noticeable performance drop occurs for known to unknown \textbf{fine motions} (red to yellow), similar to the decline in coarse accuracy (blue to green). 
\vspace{-15pt}
}
\label{fig:overall_multimodal}
\end{figure*}

\begin{figure} [t!]
\centering
\includegraphics[width=1\linewidth]{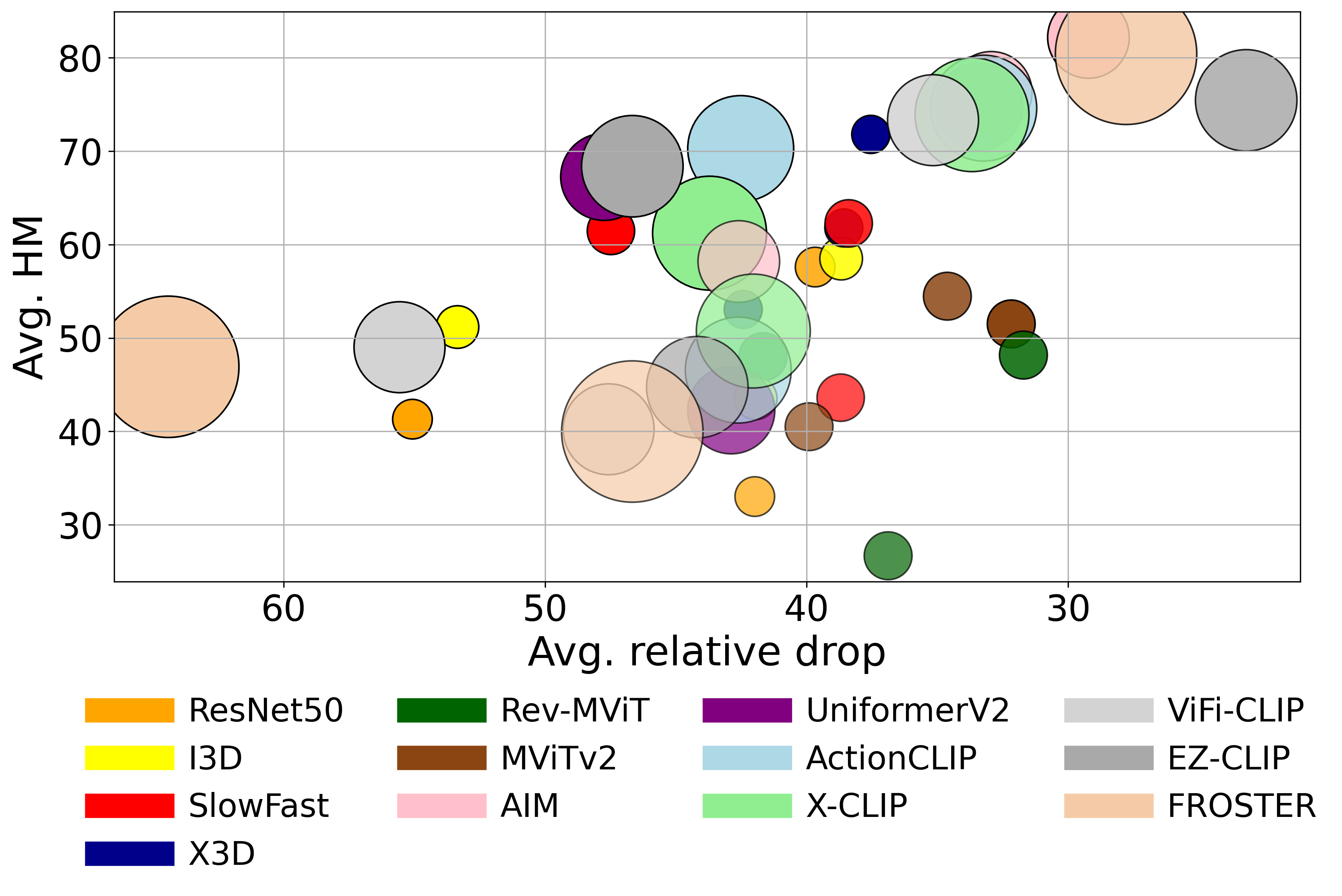}
\vspace{-20pt}
\caption{\textbf{\textit{Effect of model size on performance:}} Average harmonic mean of coarse accuracy vs. relative drop $D_{rel}$. Shape size indicates model scale, and colors distinguish specific models. 
}
 \label{fig:hm_vs_params_bubble_all}
\vspace{-10pt}
\end{figure}

\subsection{Benchmark results}
We present experimental results on motion transferability across three datasets in \cref{tab:benchmark_results_coarse} for coarse classes and \cref{tab:benchmark_results_fine} for fine classes. We report the average values of the metrics $D_{abs}$ and harmonic mean for both sets. Detailed results including $D_{rel}$ (relative drop) are provided in supplementary.

\subsection{Analysis and insights}
\paragraph{All models generalize high-level coarse classes poorly to unknown contexts.} As demonstrated in \cref{tab:benchmark_results_coarse} and \cref{fig:teaser} (right), both unimodal and multimodal models show a large drop in coarse accuracy (typical $D_{abs}$ of 20\% or more) from known to unknown contexts across all datasets. AIM performs best among all models for both Syn-TA and SSv2-TA, while being the best unimodal model for K400-TA. In K400-TA, FROSTER is the most effective (HM) when taking both types of models into account. Rev-MViT reports the lowest absolute drop in SSv2-TA and K400-TA, but it also has a low harmonic mean in those datasets, indicating that it is not detecting known classes well either. Domain generalization methods also experience drop in performance.
\vspace{-15pt}
\paragraph{Fine motions are more difficult.} For fine motions (\cref{tab:benchmark_results_fine} and \cref{fig:overall_multimodal}), the multimodal models generally have lower scores than coarse motions, but the drop in performance is still substantial, indicating that they also do not generalize well to unknown fine classes. We observe that a different model performs best for each dataset. ActionCLIP excels on Syn-TA while EZ-CLIP demonstrates well-rounded performance across SSv2-TA and K400-TA. 
\vspace{-15pt}

\begin{figure} [t!]
\centering
\includegraphics[width=1\linewidth]{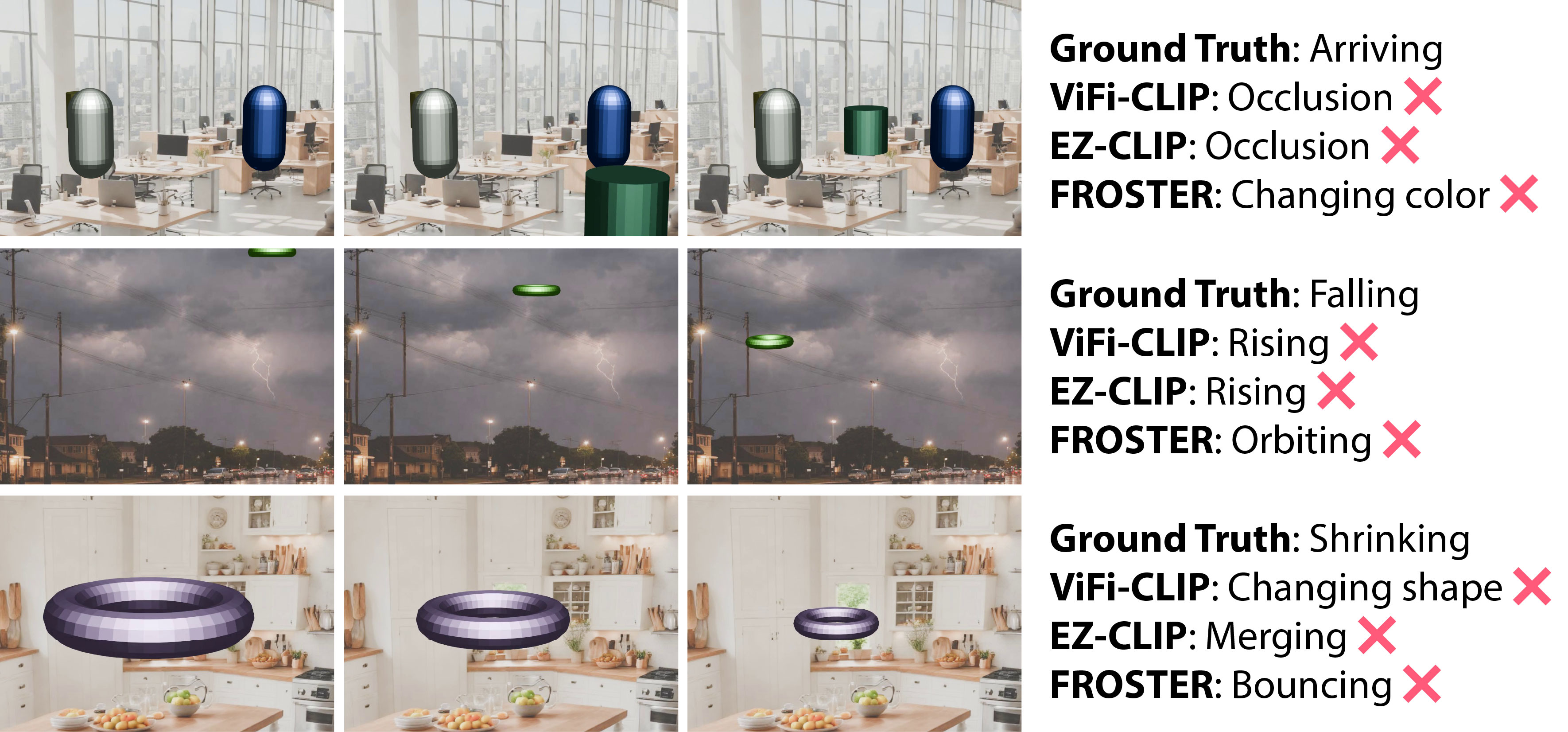}
\vspace{-15pt}
\caption{\textbf{\textit{Example of failure cases for CoarseMotion-UC in Syn-TA:}} Models misinterpret coarse motion by failing to analyze all frames, understand their sequence, or track object appearance changes, leading to motion misprediction.
}
 \label{fig:synta_model_failure_cases_head}
\vspace{-15pt}
\end{figure}

\paragraph{Focusing purely on motion (Syn-TA) is more difficult in an unknown setting without visual cues from scene (K400-TA).}
Despite its simplified setup, Syn-TA presents a greater challenge than K400-TA for most models, as indicated by the lower harmonic mean in \cref{tab:benchmark_results_coarse}. 11 out of the 13 models perform worse on Syn-TA, with four (ResNet50, X-CLIP, ViFi-CLIP, FROSTER) experiencing a drop $>$10\%. While models perform well on CoarseMotion-KC, they struggle with CoarseMotion-UC due to unseen objects and backgrounds, failing to capture temporal relationships and object-scene interactions. In \cref{fig:synta_model_failure_cases_head} (top), motion confusion arises when the green cylinder’s arrival is occluded by the blue cylinder, and in \cref{fig:synta_model_failure_cases_head} (middle), rising motion is mistaken for its opposite, falling, because the frame order is not understood. This over-reliance on familiar training contexts hinders generalization to new objects and backgrounds. This issue is less pronounced in K400-TA, where spatially dependent classes are easier to detect in unseen contexts. By controlling objects and backgrounds, Syn-TA purely tests a model’s motion understanding.
\vspace{-10pt}

\paragraph{Drop in performance from unknown coarse to fine motions is more rapid in low-bias setting (Syn-TA) than real-world videos (K400-TA, SSv2-TA).}
In \cref{fig:overall_multimodal} for known classes, we observe minimal accuracy differences (average of 4–8\%) between CoarseMotion-KC and FineMotion-K, with models performing better on coarse classes. For unknown classes, this gap varies: SSv2-TA shows a 7.9\% drop on average from CoarseMotion-UC to FineMotion-U, while K400-TA sees a slightly higher 11.14\%, with EZ-CLIP and FROSTER performing better on FineMotion-U. Syn-TA exhibits the largest gap (19.1\%) due to models struggling with object tracking, motion association, and temporal reasoning (examples in supplementary). 
This leads to fine-context mispredictions despite coarse motion understanding, along with incorrect fine-motion associations across coarse actions, resulting in Syn-TA exhibiting a greater performance drop than real-world datasets, highlighting its challenge.
\vspace{-15pt}

\begin{figure}[t!]
   \centering
   \includegraphics[width=1\linewidth]{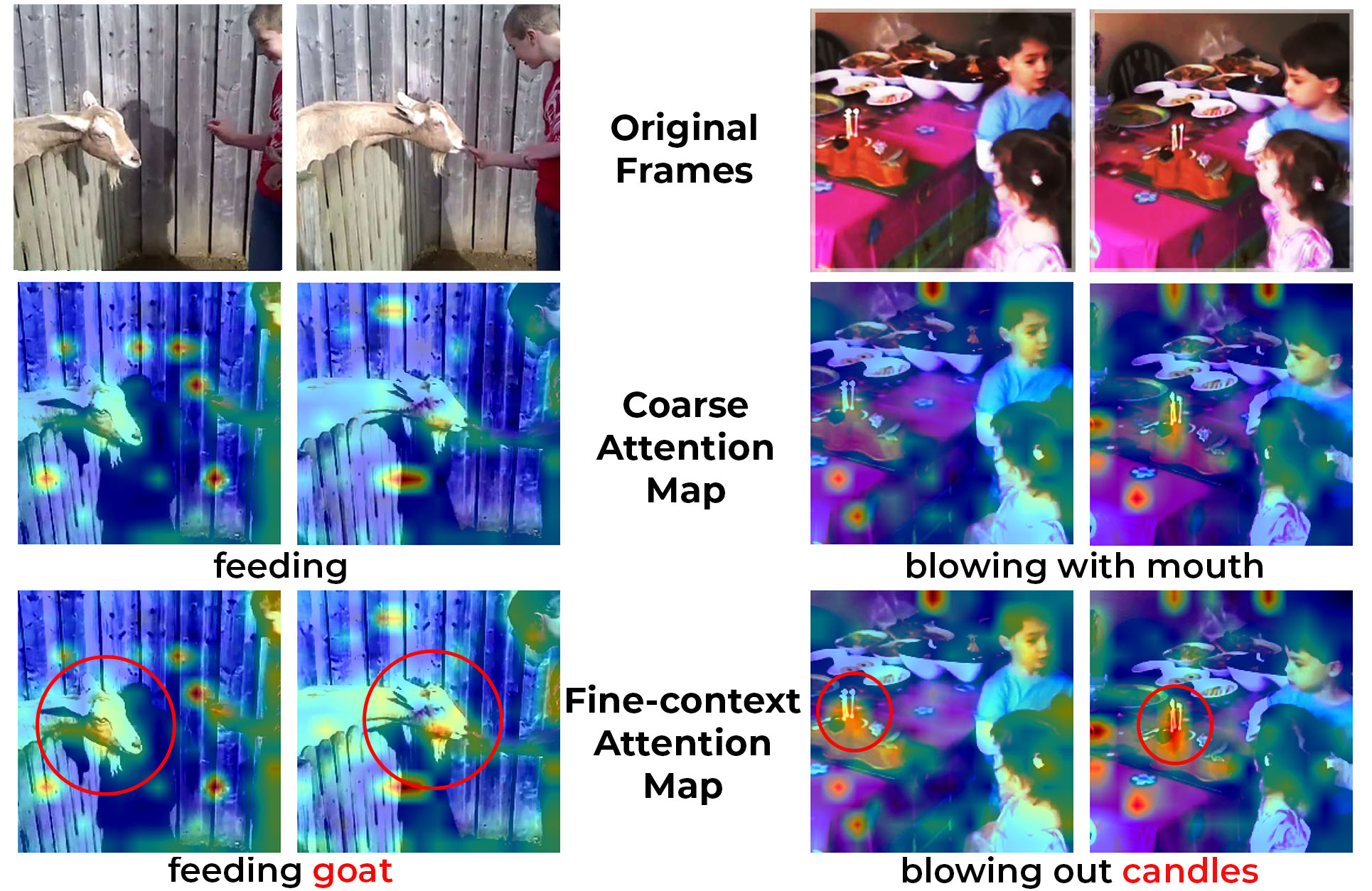}
   \vspace{-15pt}
   \caption{\textbf{\textit{K400-TA: CoarseMotion-UC vs. FineMotion-U:}} In the bottom attention map, brighter regions around `goat' or `candles' (marked by red circle) indicate that the model pays more attention there when trained with fine descriptions instead of coarse.
   } 
   \vspace{-10pt}
    \label{fig:head_vs_fine_multimodal_k400_attention}
\end{figure}

\paragraph{Poor coarse class performance doesn’t always mean poor fine-class performance.}
From previous insights, we observe that while models generally perform better on CoarseMotion-UC than FineMotion-U, EZ-CLIP and FROSTER deviate from this trend in K400-TA. To investigate, we examine EZ-CLIP’s predictions on unknown coarse classes and their fine variants in K400-TA. It struggles with unknown coarse classes ($<$30\% accuracy) but achieves over 70\% on their fine variants (detailed breakdown in supplementary). This counterintuitive behavior may stem from the model leveraging specific objects in fine captions that are absent in coarse classes. Attention maps in \cref{fig:head_vs_fine_multimodal_k400_attention} show that EZ-CLIP focuses more on objects (e.g. goat, candles) explicitly described in fine captions. These novel scenarios introduce additional object-related cues absent in coarse captions. Since both models retain CLIP weights, their pretraining biases them toward object-specific learning, leading to higher accuracy on fine classes.
\vspace{-25pt}

\begin{table}[t!]
\scriptsize
\centering
\resizebox{8cm}{!}{
\begin{tabular}{l|cc|cc}
\toprule
\textbf{Model} & \textbf{Realistic} & \textbf{Plain} & \textbf{Realistic} & \textbf{Plain} \\
\hline
\textbf{Unimodal Models} & \multicolumn{2}{c|}{\textbf{Coarse motions}} & & \\
\hline
ResNet50~\cite{resnet50} & 41.30 & \textbf{\underline{51.62}} & - & - \\
I3D~\cite{i3d} & 51.17 & \textbf{\underline{61.18}} & - & - \\
X3D~\cite{x3d} & 71.79 & \textbf{\underline{73.22}} & - & - \\
SlowFast~\cite{slowfast} & 61.45 & \textbf{\underline{72.71}} & - & - \\
MViTv2~\cite{mvitv2} & 51.50 & \textbf{\underline{61.60}} & - & - \\
Rev-MViT~\cite{rev_mvit} & 47.98 & \textbf{\underline{50.95}} & - & - \\
AIM~\cite{aim} & 82.17 & \textbf{\underline{84.02}} & - & - \\
UniformerV2~\cite{uniformerv2} & 67.25 & \textbf{\underline{75.81}} & - & - \\
\hline
\textbf{Multimodal Models} & \multicolumn{2}{c|}{\textbf{Coarse motions}} & \multicolumn{2}{c}{\textbf{Fine motions}} \\
\hline
ActionCLIP~\cite{actionclip} & 70.27 & \textbf{\underline{78.21}} & 53.85 & \textbf{\underline{56.48}} \\
X-CLIP~\cite{xclip} & 61.22 & \textbf{\underline{65.43}} & 34.98 & \textbf{\underline{36.21}} \\
ViFi-CLIP~\cite{vificlip} & 49.01 & \textbf{\underline{51.71}} & 30.79 & \textbf{\underline{37.49}} \\
EZ-CLIP~\cite{ezclip} & 68.38 & \textbf{\underline{77.17}} & 38.71 & \textbf{\underline{51.01}} \\
FROSTER~\cite{froster} & 46.91 & \textbf{\underline{59.52}} & 33.26 & \textbf{\underline{44.87}} \\
\bottomrule
\end{tabular}}
\caption{\textbf{\textit{Models perform better without the distraction of backgrounds:}} Average harmonic mean of known and unknown accuracies (averaged across two sets) for coarse and fine motions in Syn-TA with realistic vs. solid plain backgrounds.
}
\label{tab:Syn-TA-ta_realistic_vs_white}
\vspace{-20pt}
\end{table}

\begin{figure*}[t!]
\centering
{
  \includegraphics[width=1\textwidth]{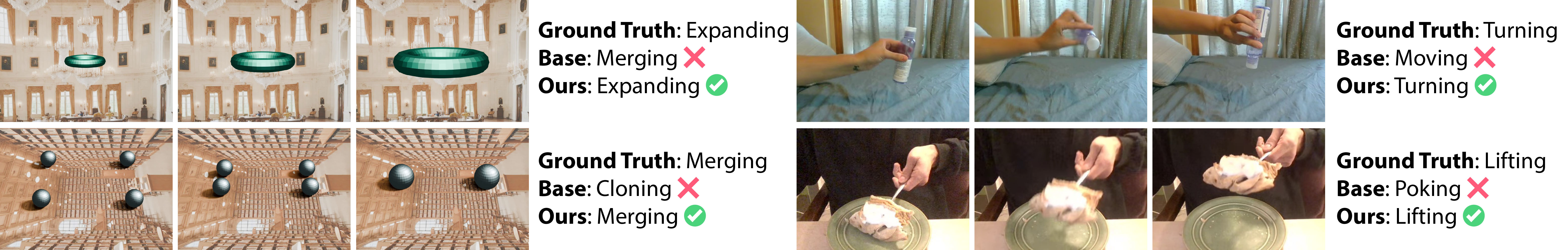}
}
\vspace{-10pt}
\caption{\textbf{\textit{Examples where our approach improves the base model:}} (left) Syn-TA: 
The base model misclassifies a single object's action as ``Merging'', failing to recognize the solitary presence. In another instance, it confuses ``Merging'' with ``Cloning'', not discerning the reversed sequence of similar frames. (right) SSv2-TA: 
The base model overlooks nuances in hand movements and object interactions, leading to misclassifications such as ``Turning'' instead of ``Moving'', and ``Poking'' instead of ``Lifting''.}
\vspace{-10pt}
\label{fig:score_improvement_base_vs_ours}
\end{figure*}

\begin{figure}[t!]
\centering
{
  \includegraphics[width=1\linewidth]{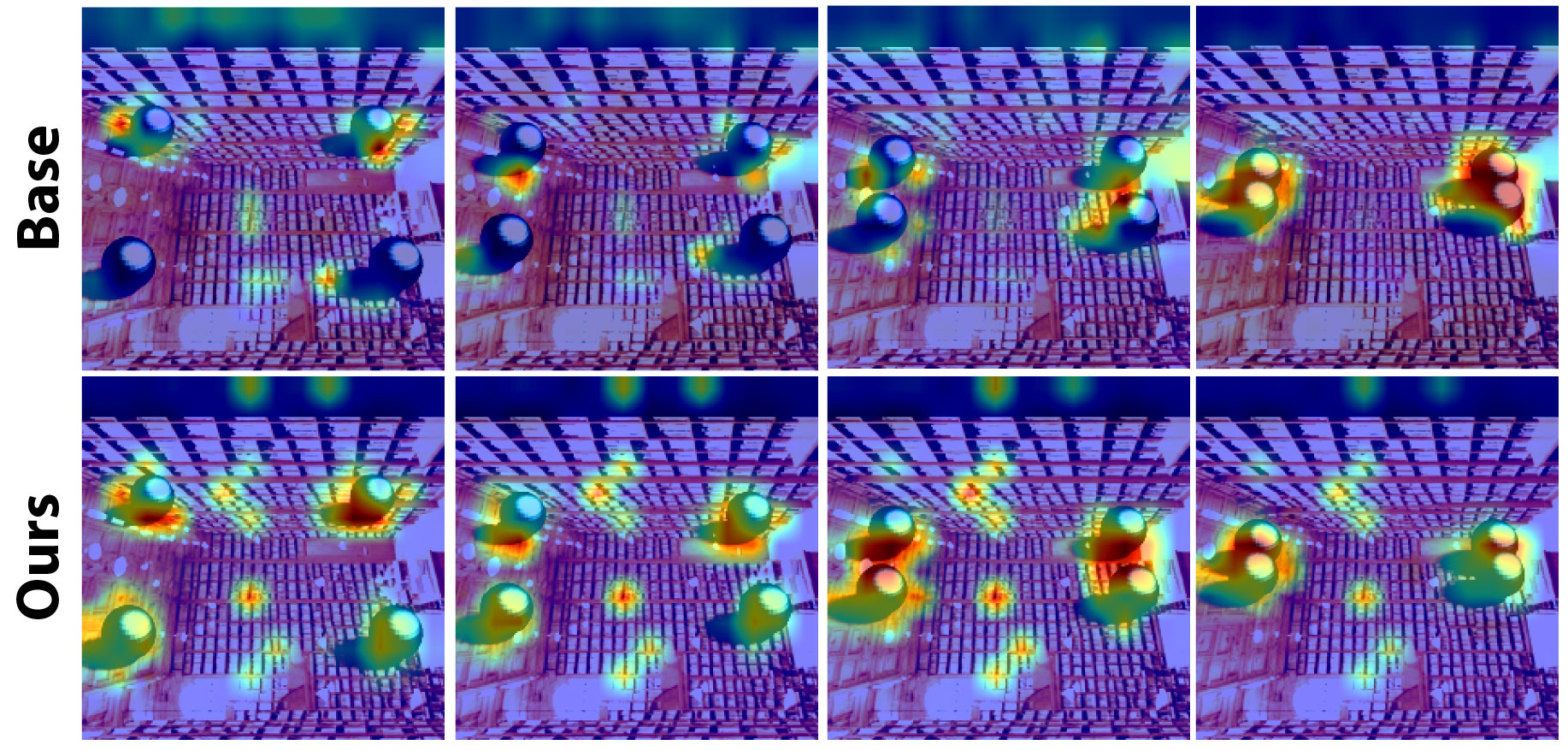}
}
\vspace{-20pt}
\caption{\textbf{\textit{A closer look at ``Merging'':}} The base model (top) ignores some objects before merging, thus mispredicting action as ``Cloning''. With disentanglement and fine cues aiding coarse detection, our approach (bottom) focuses on all objects and accurately classifies ``Merging''.
}
\vspace{-20pt}
\label{fig:score_improvement_base_vs_ours_attention}
\end{figure}

\paragraph{Larger model size improves spatial cue detection but not temporal understanding.} 
Building on our previous insights, we first calculate the average coarse performance of all models for each dataset: SSv2-TA - 43.31\%, Syn-TA - 59.26\% and K400-TA - 67.04\%. 
SSv2-TA is the most challenging, demanding strong temporal reasoning. Syn-TA follows, requiring both spatial and temporal cues from objects, background context, and motion. Models perform best on K400-TA, where classification relies primarily on spatial features. As seen in \cref{fig:hm_vs_params_bubble_all}, there is no clear correlation between model size and performance at first glance. Further looking into model performance on coarse classes for each dataset (per-dataset plots in supplementary) reveals that larger models consistently achieve better performance in K400-TA. This highlights the advantage of increased parameters in capturing spatial information. In Syn-TA, performance varies: some large models excel, while others do not, indicating that size alone does not guarantee success in mixed spatial-temporal tasks. Meanwhile, in SSv2-TA, no clear pattern emerges; larger models generally perform on par with smaller ones, indicating that parameter count does not significantly aid in understanding temporal information. Overall, while larger models excel in capturing spatial cues, this advantage does not extend to temporal understanding.
\vspace{-15pt}
\paragraph{Background texture confounds models, hindering their performance.}
To assess the impact of background information on model's understanding of motions, we created a version of Syn-TA with a solid plain background instead of realistic scenes (examples shown in supplementary). All training parameters were kept constant while we evaluated the models to measure the effect of background on activity recognition. As shown in \cref{tab:Syn-TA-ta_realistic_vs_white}, the results support our hypothesis that models rely heavily on background cues when identifying both coarse and fine motions. With reduced background information, models are forced to focus more on the motion itself rather than overfitting to irrelevant context, resulting in improved performance.
\vspace{-5pt}

\subsection{Impact of disentanglement}
\cref{tab:score_improvement_base_vs_ours} and \cref{fig:score_improvement_base_vs_ours} show how our disentanglement approach reliably corrects the base model's mistakes. In Syn-TA, the base model often misidentifies coarse actions due to inadequate comprehension of fine scene details. It mistakes object-size changes as ``Merging'' instead of ``Expanding'', possibly confusing the unfamiliar object with size changes characteristic of some ``Merging'' classes. In another case, it fails to grasp temporal order confusing ``Merging'' (multiple objects becoming one) with its inverse action, ``Cloning'' (a single object duplicating into multiple). As seen in \cref{fig:score_improvement_base_vs_ours_attention}, our model properly understands the objects involved in the ``Merging'' and correctly classifies it. In SSv2-TA, the base model confuses the motion of ``Turning'' with ``Moving'', and ``Poking'' with ``Lifting'', failing to discern the nuanced movement details of the hand. Our modified model overcomes these challenges by better understanding scene details, aided by cues from the fine branch.

\vspace{-5pt}

\begin{table}[t!]
\small
\centering
\resizebox{7cm}{!}{
\begin{tabular}{lc|c|c|c}
\toprule
\multirow{2}{*}{\textbf{Set}} & \multirow{2}{*}{\textbf{Model}} & \textbf{Syn-TA} & \textbf{K400-TA} & \textbf{SSv2-TA} \\
\cline{3-5}
& & \textbf{HM} \textuparrow & \textbf{HM} \textuparrow & \textbf{HM} \textuparrow \\
\hline
\multirow{2}{*}{\textbf{Set 1}} & Base-Coarse & 67.96 & \textbf{\underline{76.07}} & 44.83 \\
& Ours & \textbf{\underline{70.69}} & 76.06 & \textbf{\underline{46.89}} \\
\hline
\multirow{2}{*}{\textbf{Set 2}} & Base-Coarse & 68.80 & 74.80 & 44.61 \\
& Ours & \textbf{\underline{72.03}} & \textbf{\underline{75.20}} & \textbf{\underline{47.82}} \\
\bottomrule
\end{tabular}}
\caption{\textbf{\textit{Effect of disentanglement:}} Comparison of baseline vs. our proposed method for coarse motions.}
\label{tab:score_improvement_base_vs_ours}
\vspace{-20pt}
\end{table}

\section{Conclusion}
\label{sec:conclusion}
\vspace{-5pt}

We study the generalization of action recognition models across varying contexts, revealing a persistent gap in transferring high-level action knowledge to unseen fine-context actions. Syn-TA proves as challenging as real-world datasets like K400-TA, while controlled settings confirm that models rely heavily on object and background cues, limiting their generalization. We show that disentangling coarse and fine actions improves recognition, particularly in temporal datasets like Syn-TA and SSv2-TA. Our work is intended to provide a systematic benchmark for motion transferability. 
\FloatBarrier

{
    \small
    \bibliographystyle{ieeenat_fullname}
    \bibliography{main}
}
\clearpage
\maketitlesupplementary
\renewcommand{\thesection}{\Alph{section}}
\renewcommand{\thesubsection}{\thesection.\arabic{subsection}}
\setcounter{section}{0}
\setcounter{page}{1}
\maketitle
The supplementary material provides additional information to complement the main paper.
\begin{itemize}
    \item \cref{supp_models} includes a comprehensive description of the unimodal and multimodal models used in our experiments, along with implementation details. 
    \item \cref{supp_benchmark} presents detailed performance tables for models on known and unknown classes, along with more examples, failure cases and analysis figures.
    \item \cref{supp_disentanglement} shows the architecture of our proposed approach.
    \item \cref{supp_classes} provides the complete lists of coarse and fine classes for the Syn-TA, SSv2-TA, and K400-TA datasets.
    \item Datasets and relevant code are available at: \href{https://github.com/raiyaan-abdullah/Motion-Transfer}{https://github.com/raiyaan-abdullah/Motion-Transfer}.
\end{itemize}

\section{Models}
\label{supp_models}
We experimented with several unimodal and multimodal models. This includes traditional convolutional neural networks such as ResNet50 \cite{resnet50}, I3D \cite{i3d} and X3D \cite{x3d}. We also experimented with SlowFast \cite{slowfast} which utilizes slow and fast pathways. Then we explored models based on the Vision Transformer \cite{vit} such as MViTv2 \cite{mvitv2} and Rev-MViT \cite{rev_mvit} which combine multiscale features with the transformer architecture. UniformerV2 \cite{uniformerv2} combines pretrained ViTs with efficient Uniformer \cite{uniformer} designs. AIM \cite{aim} utilizes the the frozen parameters of pre-trained image models and trains various adapters. Among multimodal models, we experimented with different variations of CLIP \cite{clip} designed for activity recognition. ActionCLIP \cite{actionclip} adapts a ``pre-train, prompt, and fine-tune'' approach. X-CLIP \cite{xclip} proposes a cross-frame module and a video specific prompting scheme to adapt pre-trained language image models. ViFi-CLIP \cite{vificlip} shows that simple fine-tuning can achieve similar results to using specific temporal components. EZ-CLIP \cite{ezclip} uses temporal visual prompting and spatial adapters to efficiently prepare CLIP for downstream tasks while keeping original model weights frozen. FROSTER \cite{froster} utilizes the frozen CLIP model as a teacher for adapting to activity recognition using residual feature distillation. We also experimented with domain generalization methods such as VideoDG \cite{videodg}, STDN \cite{stdn}, and CIR \cite{cookitalyindia} on coarse classes.

\textbf{Implementation details:}
For training ResNet50, I3D, X3D, MViTv2, Rev-MViT, and SlowFast we utilized the \href{https://github.com/facebookresearch/SlowFast}{PySlowFast repository} \cite{pyslowfast} from Meta Research. For other models and domain generalization methods, we used the code from their respective GitHub repositories. We particularly used the model versions: I3D R50, X3D-M, SlowFast R50, MViTv2-S, Rev-MViT-B-16, AIM ViT-B/16, UniFormerV2-B/16, ActionCLIP ViT-B/16, X-CLIP-B/16, EZ-CLIP ViT-B/16, and FROSTER-B/16. The model hyperparameters for training were kept similar to their configuration for Something-something-v2 and Kinetics400. For Syn-TA, we followed the respective hyperparameters of Kinetics400 for each model. The learning rate was slightly tuned in some cases. We also modified the configurations dependent on the compute machine like batch size, number of GPUs, number of workers, etc to adjust to our resources. The number of epochs was varied by model and dataset depending on how fast the model converges. The models were trained on 1-4 NVIDIA GPUs. The memory of GPUs varied from 11 GB to 80 GB. The configuration files for training the models are available in our \href{https://github.com/raiyaan-abdullah/Motion-Transfer}{GitHub}.

\section{Benchmark results}
\begin{figure*}[h!]
   \centering
   \includegraphics[width=1\textwidth]{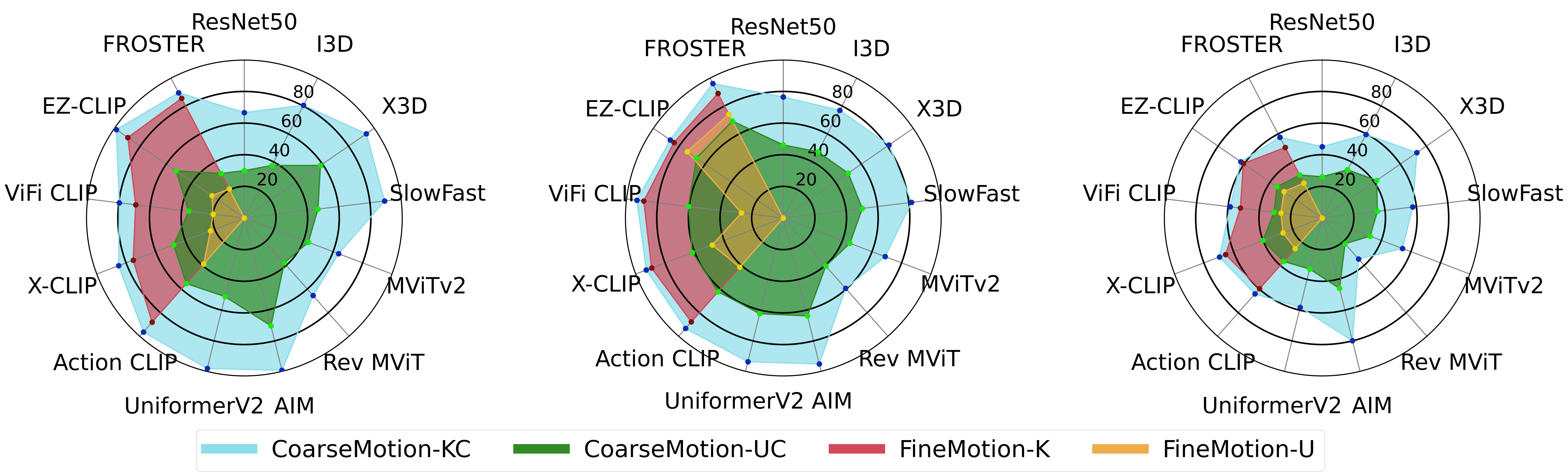}
   \caption{Left: Syn-TA, middle: K400-TA, and right: SSv2-TA. Average detection scores of both sets are given for the three datasets. Performance drop is observed for both coarse (blue to green for all models) and fine motions (red to yellow for multimodal models).}
   
    \label{fig:overall}
\end{figure*}

Along with $D_{abs}$ and HM, we show an additional metric:\\
\textbf{Relative Drop}: The percentage decrease in performance when shifting to an unknown context:
$$D_{rel} = |\frac{Known - Unknown}{Known}| \times 100$$
\vspace{-5pt}
\label{supp_benchmark}
\paragraph{Performance on known vs. unknown classes:} The known and unknown accuracies for both Set 1 and Set 2, covering coarse and fine classes, along with other metrics, are detailed in \cref{tab:synthetic_results_head_detailed}, \cref{tab:synthetic_results_fine_detailed} (Syn-TA); \cref{tab:k400_results_head_detailed}, \cref{tab:k400_results_fine_detailed} (K400-TA), and \cref{tab:ssv2_results_head_detailed}, \cref{tab:ssv2_results_fine_detailed} (SSv2-TA). For metrics such as known accuracy, unknown accuracy, and harmonic mean (HM), higher values indicate better performance, whereas lower values are desirable for $D_{abs}$ and $D_{rel}$. 
As discussed in the main paper, there is a noticeable drop in performance for both coarse and fine motions across all models, illustrated more clearly in \cref{fig:overall}.

\paragraph{Performance on coarse vs fine classes:} \cref{fig:overall} also shows that fine classes are generally more challenging than coarse classes. However, notable exceptions include the performance of EZ-CLIP and FROSTER in K400-TA unknown classes. 

\begin{table*}[h!]
\centering
\resizebox{16cm}{!}{%
\begin{tabular}{l|ccccc|ccccc}
\toprule
\multirow{2}{*}{\textbf{Model}} & \multicolumn{5}{c}{\textbf{Set 1}} & \multicolumn{5}{c}{\textbf{Set 2}} \\
\cline{2-11}
{} & \textbf{Known (Set 1)} \textuparrow & \textbf{Unknown (Set 2)} \textuparrow & \textbf{D\textsubscript{abs}} \textdownarrow & \textbf{D\textsubscript{rel}} \textdownarrow & \textbf{HM} \textuparrow & \textbf{Known (Set 2)} \textuparrow & \textbf{Unknown (Set 1)} \textuparrow & \textbf{D\textsubscript{abs}} \textdownarrow & \textbf{D\textsubscript{rel}} \textdownarrow & \textbf{HM} \textuparrow \\
\hline
{} & \multicolumn{10}{c}{\textbf{Unimodal models}} \\
\hline
ResNet50~\cite{resnet50} & 67.59 & 29.26 & 38.33 & 56.70 & 40.84 & 65.74 & 30.61 & 35.13 & 53.43 & 41.77 \\
I3D~\cite{i3d} & 83.40 & 37.77 & 45.63 & 54.71 & 51.99 & 77.61 & 37.26 & 40.35 & 51.99 & 50.34 \\
X3D~\cite{x3d} & 94.76 & 52.71 & 42.05 & 44.37 & 67.74 & 92.66 & 64.20 & 28.46 & 30.71 & 75.84 \\
SlowFast~\cite{slowfast} & 90.94 & 46.70 & 44.24 & 48.64 & 61.71 & 87.61 & 47.03 & 40.58 & 46.31 & 61.20 \\
MViTv2~\cite{mvitv2} & 57.50 & 38.72 & \textbf{\underline{18.78}} & 32.66 & 46.27 & 69.89 & 47.74 & \textbf{\underline{22.15}} & 31.69 & 56.73 \\
Rev-MViT~\cite{rev_mvit} & 62.45 & 40.48 & 21.97 & 35.18 & 49.12 & 68.62 & 35.57 & 33.05 & 48.16 & 46.85 \\
AIM~\cite{aim} & \textbf{\underline{99.81}} & \textbf{\underline{70.85}} & 28.96 & \textbf{\underline{29.01}} & \textbf{\underline{82.87}} & 98.46 & \textbf{\underline{69.48}} & 28.98 & \textbf{\underline{29.43}} & \textbf{\underline{81.47}} \\
UniformerV2~\cite{uniformerv2} & 96.93 & 50.32 & 46.61 & 48.08 & 66.24 & \textbf{\underline{98.99}} & 52.08 & 46.91 & 47.38 & 68.25 \\
\hline
{} & \multicolumn{10}{c}{\textbf{Multimodal models}} \\
\hline
ActionCLIP~\cite{actionclip} & 97.74 & 55.43 & 42.31 & 43.28 & 70.74 & 94.84 & 55.24 & 39.60 & 41.75 & 69.81 \\
X-CLIP~\cite{xclip} & 87.54 & 48.03 & 39.51 & 45.13 & 62.02 & 82.55 & 47.64 & 34.91 & 42.28 & 60.41 \\
ViFi-CLIP~\cite{vificlip} & 81.32 & 39.04 & 42.28 & 51.99 & 52.75 & 78.03 & 31.88 & 46.15 & 59.14 & 45.26 \\
EZ-CLIP~\cite{ezclip} & 98.38 & 51.92 & 46.46 & 47.22 & 67.96 & 98.23 & 52.94 & 45.29 & 46.10 & 68.80 \\
FROSTER~\cite{froster} & 91.13 & 31.44 & 59.69 & 65.49 & 46.75 & 87.71 & 32.17 & 55.54 & 63.32 & 47.07 \\
\hline
{} & \multicolumn{10}{c}{\textbf{Domain Generalization Methods}} \\
\hline
VideoDG~\cite{videodg} & 98.34 & 45.69 & 52.65 & 53.53 & 62.39 & 97.81 & 41.17 & 56.64 & 57.90 & 57.94 \\
STDN~\cite{stdn} & 70.80 & 26.48 & 44.32 & 62.59 & 38.54 & 70.53 & 21.46 & 49.07 & 69.57 & 32.90 \\
CIR~\cite{cookitalyindia} & 62.01 & 7.42 & 54.59 & 88.03 & 13.25 & 58.25 & 11.76 & 46.49 & 79.81 & 19.56 \\
\bottomrule
\end{tabular}}
\caption{Known and unknown accuracy of coarse motions on Syn-TA}
\label{tab:synthetic_results_head_detailed}
\end{table*}

\begin{table*}[h!]
\centering
\resizebox{16cm}{!}{%
\begin{tabular}{l|ccccc|ccccc}
\toprule
\multirow{2}{*}{\textbf{Model}} & \multicolumn{5}{c}{\textbf{Set 1}} & \multicolumn{5}{c}{\textbf{Set 2}} \\
\cline{2-11}
{} & \textbf{Known (Set 1)} \textuparrow & \textbf{Unknown (Set 2)} \textuparrow & \textbf{D\textsubscript{abs}} \textdownarrow & \textbf{D\textsubscript{rel}} \textdownarrow & \textbf{HM} \textuparrow & \textbf{Known (Set 2)} \textuparrow & \textbf{Unknown (Set 1)} \textuparrow & \textbf{D\textsubscript{abs}} \textdownarrow & \textbf{D\textsubscript{rel}} \textdownarrow & \textbf{HM} \textuparrow \\
\hline
ActionCLIP~\cite{actionclip} & 89.58 & \textbf{\underline{40.74}} & 48.84 & \textbf{\underline{54.52}} & \textbf{\underline{56.00}} & 86.44 & \textbf{\underline{36.89}} & \textbf{\underline{49.55}} & \textbf{\underline{57.32}} & \textbf{\underline{51.71}} \\
X-CLIP~\cite{xclip} & 75.99 & 26.43 & 49.56 & 65.21 & 39.21 & 74.41 & 19.38 & 55.03 & 73.95 & 30.75 \\
ViFi-CLIP~\cite{vificlip} & 72.21 & 23.88 & \textbf{\underline{48.33}} & 66.92 & 35.89 & 66.33 & 15.94 & 50.39 & 75.96 & 25.70 \\
EZ-CLIP~\cite{ezclip} & \textbf{\underline{91.69}} & 30.23 & 61.46 & 67.03 & 45.46 & \textbf{\underline{87.39}} & 19.56 & 67.83 & 77.61 & 31.96 \\
FROSTER~\cite{froster} & 87.26 & 22.82 & 64.44 & 73.84 & 36.17 & 83.62 & 18.54 & 65.08 & 77.82 & 30.35 \\
\bottomrule
\end{tabular}}
\caption{Known and unknown accuracy of fine motions on Syn-TA}
\label{tab:synthetic_results_fine_detailed}
\end{table*}

\begin{table*}[h!]
\centering
\resizebox{16cm}{!}{%
\begin{tabular}{l|ccccc|ccccc}
\toprule
\multirow{2}{*}{\textbf{Model}} & \multicolumn{5}{c}{\textbf{Set 1}} & \multicolumn{5}{c}{\textbf{Set 2}} \\
\cline{2-11}
{} & \textbf{Known (Set 1)} \textuparrow & \textbf{Unknown (Set 2)} \textuparrow & \textbf{D\textsubscript{abs}} \textdownarrow & \textbf{D\textsubscript{rel}} \textdownarrow & \textbf{HM} \textuparrow & \textbf{Known (Set 2)} \textuparrow & \textbf{Unknown (Set 1)} \textuparrow & \textbf{D\textsubscript{abs}} \textdownarrow & \textbf{D\textsubscript{rel}} \textdownarrow & \textbf{HM} \textuparrow \\
\hline
{} & \multicolumn{10}{c}{\textbf{Unimodal models}} \\
\hline
ResNet50~\cite{resnet50} & 79.27 & 49.74 & 29.53 & 37.25 & 61.12 & 73.71 & 42.68 & 31.03 & 42.09 & 54.05 \\
I3D~\cite{i3d} & 80.09 & 51.76 & 28.33 & 35.37 & 62.88 & 73.69 & 42.75 & 30.94 & 41.98 & 54.10 \\
X3D~\cite{x3d} & 80.66 & 51.42 & 29.24 & 36.25 & 62.80 & 81.80  & 48.34 & 33.46 & 40.90 & 60.76 \\
SlowFast~\cite{slowfast} & 81.61 & 52.40  & 29.21 & 35.79 & 63.82 & 81.80  & 48.27 & 33.53 & 40.99 & 60.71 \\
MViTv2~\cite{mvitv2} & 70.46 & 47.41 & 23.05 & 32.71 & 56.68 & 67.30  & 42.72 & 24.58 & 36.52 & 52.26 \\
Rev-MViT~\cite{rev_mvit} & 58.62 & 41.80 & 16.82 & 28.69 & 48.80 & 60.18 & 39.28 & \textbf{\underline{20.90}} & 34.72 & 47.53 \\
AIM~\cite{aim} & \textbf{\underline{95.00}}  & 64.77 & 30.23 & 31.82 & 77.02 & 95.09 & 62.70 & 32.39 & 34.06 & 75.57 \\
UniformerV2~\cite{uniformerv2} & 92.81 & 63.81 & 29.00 & 31.24 & 75.62 & 94.32 & 60.77 & 33.55 & 35.57 & 73.91 \\
\hline
{} & \multicolumn{10}{c}{\textbf{Multimodal models}} \\
\hline
ActionCLIP~\cite{actionclip} & 92.75 & 65.20 & 27.55 & 29.70 & 76.57 & 93.74 & 59.29 & 34.45 & 36.75 & 72.63 \\
X-CLIP~\cite{xclip} & 92.54 &  63.36 & 29.18 & 31.53 & 75.21 & 92.85 & 59.59 & 33.26 & 35.82 & 72.59 \\
ViFi-CLIP~\cite{vificlip} & 92.66 & 62.30 & 30.36 & 32.76 & 74.50 & 93.83 & 58.58 & 35.25 & 37.56 & 72.12 \\
EZ-CLIP~\cite{ezclip} & 85.70 & 68.39 & \textbf{\underline{17.31}} & \textbf{\underline{20.19}} & 76.07 & 88.07 & 65.01 & 23.06 & \textbf{\underline{26.18}} & 74.80 \\
FROSTER~\cite{froster} & 92.68 & \textbf{\underline{69.38}} & 23.30 & 25.14  & \textbf{\underline{79.35}} & \textbf{\underline{99.31}} & \textbf{\underline{69.09}} & 30.22 & 30.43 & \textbf{\underline{81.48}} \\
\hline
{} & \multicolumn{10}{c}{\textbf{Domain Generalization Methods}} \\
\hline
VideoDG~\cite{videodg} & 84.94 & 56.63 & 28.31 & 33.32 & 67.95 & 87.28 & 51.28 & 36.00 & 41.24 & 64.60 \\
STDN~\cite{stdn} & 65.17 & 47.96 & 17.21 & 26.40 & 55.25 & 71.05 & 44.24 & 26.81 & 37.73 & 54.52 \\
CIR~\cite{cookitalyindia} & 68.53 & 13.82 & 54.71 & 79.83 & 23.00 & 47.22 & 11.50 & 57.03 & 83.21 & 19.69 \\
\bottomrule
\end{tabular}}
\caption{Known and unknown accuracy of coarse motions on K400-TA}
\label{tab:k400_results_head_detailed}
\end{table*}

\begin{table*}[h!]
\centering
\resizebox{16cm}{!}{%
\begin{tabular}{l|ccccc|ccccc}
\toprule
\multirow{2}{*}{\textbf{Model}} & \multicolumn{5}{c}{\textbf{Set 1}} & \multicolumn{5}{c}{\textbf{Set 2}} \\
\cline{2-11}
{} & \textbf{Known (Set 1)} \textuparrow & \textbf{Unknown (Set 2)} \textuparrow & \textbf{D\textsubscript{abs}} \textdownarrow & \textbf{D\textsubscript{rel}} \textdownarrow & \textbf{HM} \textuparrow & \textbf{Known (Set 2)} \textuparrow & \textbf{Unknown (Set 1)} \textuparrow & \textbf{D\textsubscript{abs}} \textdownarrow & \textbf{D\textsubscript{rel}} \textdownarrow & \textbf{HM} \textuparrow \\
\hline
ActionCLIP~\cite{actionclip} & 86.84 & 45.86 & 40.98 & 47.19 & 60.02 & 88.66 & 37.18 & 51.48 & 58.06 & 52.39 \\
X-CLIP~\cite{xclip} & \textbf{\underline{88.98}} & 51.95 & 37.03 & 41.61 & 65.60 & 89.14 & 44.27 & 44.87 &  50.33 & 59.15 \\
ViFi-CLIP~\cite{vificlip} & 88.44 & 29.63 & 58.81 & 66.49 & 44.38 & 89.38 & 23.77 & 65.61 & 73.40 & 37.55 \\
EZ-CLIP~\cite{ezclip} & 82.24 & \textbf{\underline{77.20}} & \textbf{\underline{5.04}} & \textbf{\underline{6.12}} & 79.64 & 85.29 & 70.70 & \textbf{\underline{14.59}} & \textbf{\underline{17.10}} & 77.31 \\
FROSTER~\cite{froster} & 88.35 & 76.97 & 11.38 & 12.88 & \textbf{\underline{82.26}} & \textbf{\underline{89.52}} & \textbf{\underline{71.26}} & 18.26 & 20.39 & \textbf{\underline{79.35}} \\
\bottomrule
\end{tabular}}
\caption{Known and unknown accuracy of fine motions on K400-TA}
\label{tab:k400_results_fine_detailed}
\end{table*}

\begin{table*}[h!]
\centering
\resizebox{16cm}{!}{%
\begin{tabular}{l|ccccc|ccccc}
\toprule
\multirow{2}{*}{\textbf{Model}} & \multicolumn{5}{c}{\textbf{Set 1}} & \multicolumn{5}{c}{\textbf{Set 2}} \\
\cline{2-11}
{} & \textbf{Known (Set 1)} \textuparrow & \textbf{Unknown (Set 2)} \textuparrow & \textbf{D\textsubscript{abs}} \textdownarrow & \textbf{D\textsubscript{rel}} \textdownarrow & \textbf{HM} \textuparrow & \textbf{Known (Set 2)} \textuparrow & \textbf{Unknown (Set 1)} \textuparrow & \textbf{D\textsubscript{abs}} \textdownarrow & \textbf{D\textsubscript{rel}} \textdownarrow & \textbf{HM} \textuparrow \\
\hline
{} & \multicolumn{10}{c}{\textbf{Unimodal models}} \\
\hline
ResNet50~\cite{resnet50} & 47.02  & 25.67  & 21.35 & 45.40 & 33.21  & 43.13  & 26.50  & 16.63 & 38.55 & 32.82 \\
I3D~\cite{i3d} & 63.19  & 33.20   & 29.99 & 47.46 & 43.53  & 56.02  & 35.61  & 20.41 & 36.43 & 43.54 \\
X3D~\cite{x3d} & 74.43  & 40.11  & 34.32 & 46.11  & 52.12 & 71.04  & 43.52  & 27.52 & 38.73 & 53.97 \\
SlowFast~\cite{slowfast} & 61.47  & 34.48  & 26.99  & 43.90 & 44.17 & 53.87  & 35.83  & 18.04 & 33.48 & 43.03 \\
MViTv2~\cite{mvitv2} & 59.46  & 32.71  & 26.75 & 44.98 & 42.20 & 49.16  & 32.04  & 17.12 & 34.82 & 38.79 \\
Rev-MViT~\cite{rev_mvit} & 38.32  & 22.71  & \textbf{\underline{15.61}} & \textbf{\underline{40.73}} & 28.51 & 30.97  & 20.74  & \textbf{\underline{10.23}} & \textbf{\underline{33.03}} & 24.84 \\
AIM~\cite{aim} & \textbf{\underline{81.62}}  & \textbf{\underline{43.52}}  & 38.10 & 46.67 & \textbf{\underline{56.77}}  & \textbf{\underline{78.27}}  & \textbf{\underline{48.13}}  & 30.14 & 38.50 & \textbf{\underline{59.60}} \\
UniformerV2~\cite{uniformerv2} & 59.06  & 32.47  & 26.59  & 45.02 & 41.90 & 57.26  & 33.93  & 23.33 & 40.74 & 42.61 \\
\hline
{} & \multicolumn{10}{c}{\textbf{Multimodal models}} \\
\hline
ActionCLIP~\cite{actionclip} & 66.44 & 34.67  & 31.77 & 47.81 & 45.56 & 61.77 & 38.66  & 23.11 & 37.41 & 47.55 \\
X-CLIP~\cite{xclip} & 72.50 & 37.80 & 34.70 & 47.86 & 49.69 & 66.49  & 42.41 & 24.08 & 36.21 & 51.78 \\
ViFi-CLIP~\cite{vificlip} & 60.17  & 28.49 & 31.68 & 52.65 & 38.67 & 57.21 & 32.90 & 24.31 & 42.49 & 41.77 \\
EZ-CLIP~\cite{ezclip} & 64.86  & 34.26 & 30.60 & 47.17 & 44.83 & 60.24  & 35.43  & 24.81 & 41.18 & 44.61 \\
FROSTER~\cite{froster} & 59.01 & 28.68 & 30.33 & 51.39 & 38.60 & 56.30 & 32.69 & 23.61 & 41.93 & 41.36 \\
\hline
{} & \multicolumn{10}{c}{\textbf{Domain Generalization Methods}} \\
\hline
VideoDG~\cite{videodg} & 59.21 & 30.69 & 28.52 & 48.16 & 40.42 & 55.30 & 32.39 & 22.91 & 41.42 & 40.85 \\
STDN~\cite{stdn} & 37.44 & 22.33 & 15.11 & 40.35 & 27.97 & 34.43 & 22.29 & 12.14 & 35.25 & 27.06 \\
CIR~\cite{cookitalyindia} & 48.80 & 31.84 & 16.96 & 34.75 & 38.53 & 47.22 & 32.10 & 15.12 & 32.02 & 38.21 \\
\bottomrule
\end{tabular}}
\caption{Known and unknown accuracy of coarse motions on SSv2-TA}
\label{tab:ssv2_results_head_detailed}
\end{table*}

\begin{table*}[h!]
\centering
\resizebox{16cm}{!}{%
\begin{tabular}{l|ccccc|ccccc}
\toprule
\multirow{2}{*}{\textbf{Model}} & \multicolumn{5}{c}{\textbf{Set 1}} & \multicolumn{5}{c}{\textbf{Set 2}} \\
\cline{2-11}
{} & \textbf{Known (Set 1)} \textuparrow & \textbf{Unknown (Set 2)} \textuparrow & \textbf{D\textsubscript{abs}} \textdownarrow & \textbf{D\textsubscript{rel}} \textdownarrow & \textbf{HM} \textuparrow & \textbf{Known (Set 2)} \textuparrow & \textbf{Unknown (Set 1)} \textuparrow & \textbf{D\textsubscript{abs}} \textdownarrow & \textbf{D\textsubscript{rel}} \textdownarrow & \textbf{HM} \textuparrow \\
\hline
ActionCLIP~\cite{actionclip} & 60.12  & 24.10   & 36.02 & 59.91 & 34.40 & 59.33  & 27.58  & 31.75 & 53.51 & 37.65 \\
X-CLIP~\cite{xclip} & \textbf{\underline{66.01}}  & 24.92  & 41.09  & 62.24 & 36.18 & \textbf{\underline{64.61}}  & \textbf{\underline{28.14}}  & 36.47 & 56.44 & \textbf{\underline{39.20}} \\
ViFi-CLIP~\cite{vificlip} & 54.02 & 26.59 & \textbf{\underline{27.43}} & 50.77 & 35.63  & 50.24 & 25.97 & 24.27 & 48.30 & 34.24  \\
EZ-CLIP~\cite{ezclip} & 61.08  & \textbf{\underline{31.83}} & 29.25 & \textbf{\underline{47.88}} & \textbf{\underline{41.85}} & 58.58  & 27.64  & 30.94 & 52.81 & 37.55 \\
FROSTER~\cite{froster} & 51.90   & 23.58  & 28.32 & 54.56 & 32.42 & 48.79  & 26.40 & \textbf{\underline{22.39}} & \textbf{\underline{45.89}} & 34.26 \\
\bottomrule
\end{tabular}}
\caption{Known and unknown accuracy of fine motions on SSv2-TA}
\label{tab:ssv2_results_fine_detailed}
\end{table*}

\begin{figure*} [h!]
\centering
\includegraphics[width=1\linewidth]{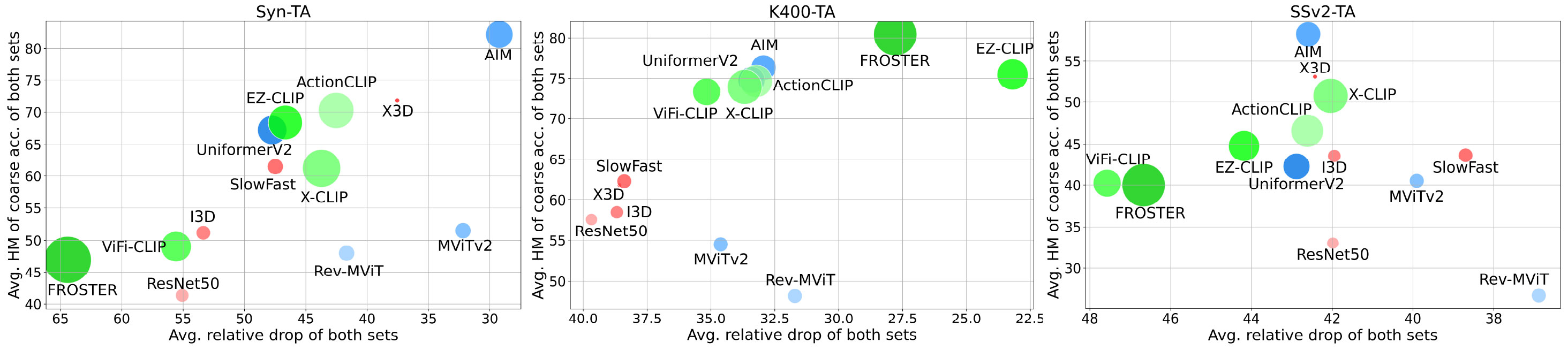}
\vspace{-10pt}
\caption{\textbf{\textit{Effect of model size for each dataset:}} Average harmonic mean of coarse accuracy vs. relative drop $D_{rel}$. Bubble sizes correspond to the total number of model parameters, with colors indicating architecture types (red: CNN, blue: transformer - unimodal, green: transformer - multimodal). Models with larger parameter counts perform better on K400-TA, where videos contain rich spatial cues. This effect is less pronounced in Syn-TA, which requires some temporal understanding. In SSv2-TA, which is heavily reliant on temporal information, model size does not show a clear correlation with performance.
}
 \label{fig:hm_vs_params_bubble}
\vspace{-30pt}
\end{figure*}

\begin{figure*}[h!]
\centering
\includegraphics[width=0.8\textwidth]{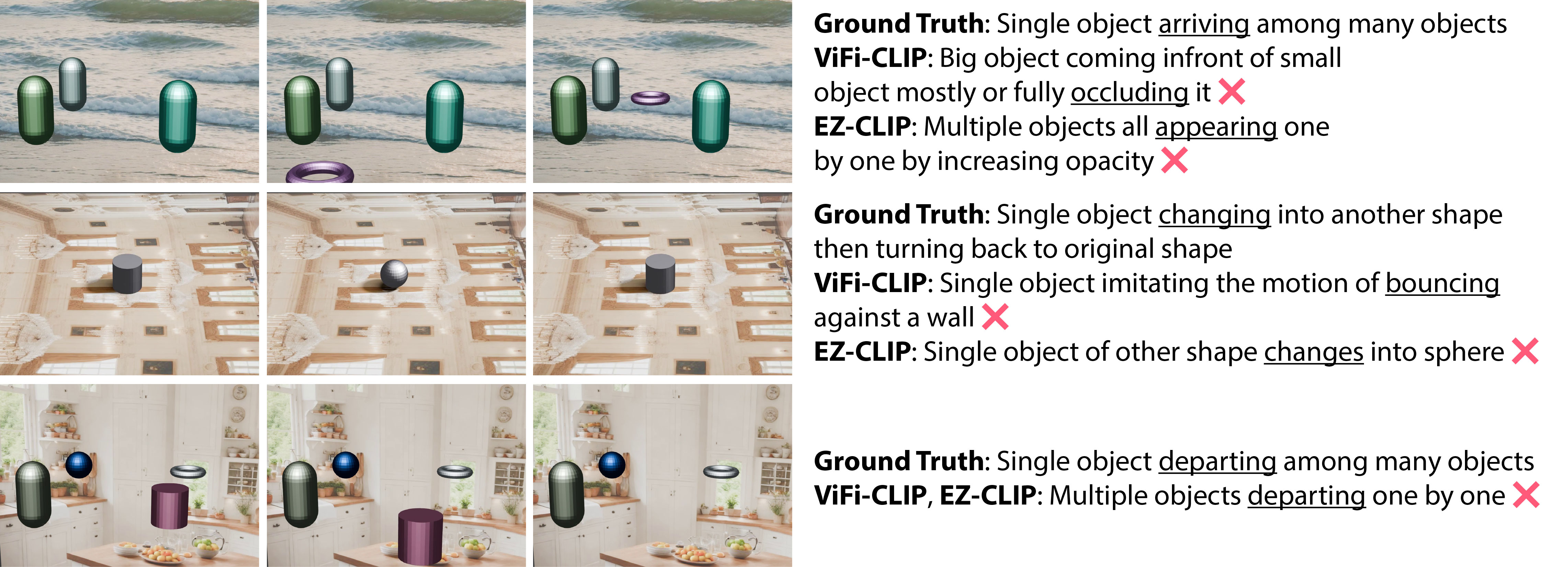}
\caption{\textbf{\textit{Example of failure cases for FineMotion-U in Syn-TA with ViFi-CLIP, EZ-CLIP:}} (Top) For the arrival of single objects among many objects, ViFi-CLIP is confusing the motion with occlusion as the arriving pink torus temporarily occludes the objects before completing its path. EZ-CLIP hallucinates the multiple objects are appearing in the scene. (Middle) The object in the scene changes its shape to a sphere and turns back. EZ-CLIP thinks the object did not transform back. (Bottom) Both models are mispredicting that the other objects are departing as well after the first object.
}
\vspace{-30pt}
 \label{fig:synta_model_failure_cases_fine}
\end{figure*}

\begin{figure*}[!t]
\centering
{
  \includegraphics[width=1\textwidth]{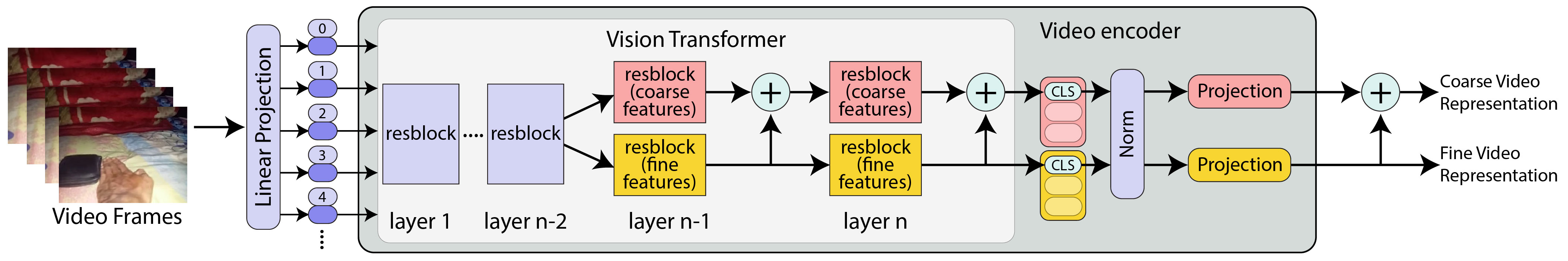}
}
\vspace{-10pt}
\caption{\textbf{\textit{Disentanglement of coarse and fine video features:}} In the final two layers of the vision transformer, two branches extract coarse and fine motions simultaneously. The fine embeddings are added to the high-level embedding at each step via residual connections, combining detailed context with the broader motion features. The fine projection layer is trainable. Overall, this enables each branch to focus on disentangling the features most relevant to its specific role.}
\vspace{-10pt}
\label{fig:archi_improvement}
\end{figure*}

\begin{table*}[]
\small
\centering
\resizebox{12cm}{!}{
\begin{tabular}{lc|cc|cc|cc}
\toprule
\multirow{2}{*}{\textbf{Set}} & \multirow{2}{*}{\textbf{Model}} & \multicolumn{2}{c}{\textbf{Syn-TA}} & \multicolumn{2}{c}{\textbf{K400-TA}} & \multicolumn{2}{c}{\textbf{SSv2-TA}} \\
\cline{3-8}
& & \textbf{Known/Unknown} & \textbf{HM} & \textbf{Known/Unknown} & \textbf{HM} & \textbf{Known/Unknown} & \textbf{HM} \\
\hline
& & \multicolumn{6}{c}{\textbf{Coarse motions}} \\
\hline
\multirow{2}{*}{\textbf{Set 1}} & Base-Coarse & 98.38/51.92 & 67.96 & \textbf{\underline{85.70}}/68.39 & \textbf{\underline{76.07}} & 64.86/34.26 & 44.83 \\
& Ours & \textbf{\underline{99.76}}/\textbf{\underline{54.75}} & \textbf{\underline{70.69}} & 84.15/\textbf{\underline{69.39}} & 76.06 & \textbf{\underline{69.17}}/\textbf{\underline{35.47}} & \textbf{\underline{46.89}} \\
\hline
\multirow{2}{*}{\textbf{Set 2}} & Base-Coarse & 98.23/52.94 & 68.80 & 88.07/65.01 & 74.80 & 60.24/35.43 & 44.61 \\
& Ours & \textbf{\underline{99.62}}/\textbf{\underline{56.41}} & \textbf{\underline{72.03}} & \textbf{\underline{88.53}}/\textbf{\underline{65.36}} & \textbf{\underline{75.20}} & \textbf{\underline{65.01}}/\textbf{\underline{37.82}} & \textbf{\underline{47.82}} \\
\bottomrule
\end{tabular}}
\caption{\textbf{\textit{Performance comparison for disentanglement approach:}} Comparison of baseline vs. our proposed method for coarse motions.
}
\label{tab:score_improvement_base_vs_ours_detailed}
\vspace{-15pt}
\end{table*}

\FloatBarrier
\clearpage

\section{Disentanglement architecture}

\begin{figure}[!t]
\centering
\includegraphics[width=1\linewidth]{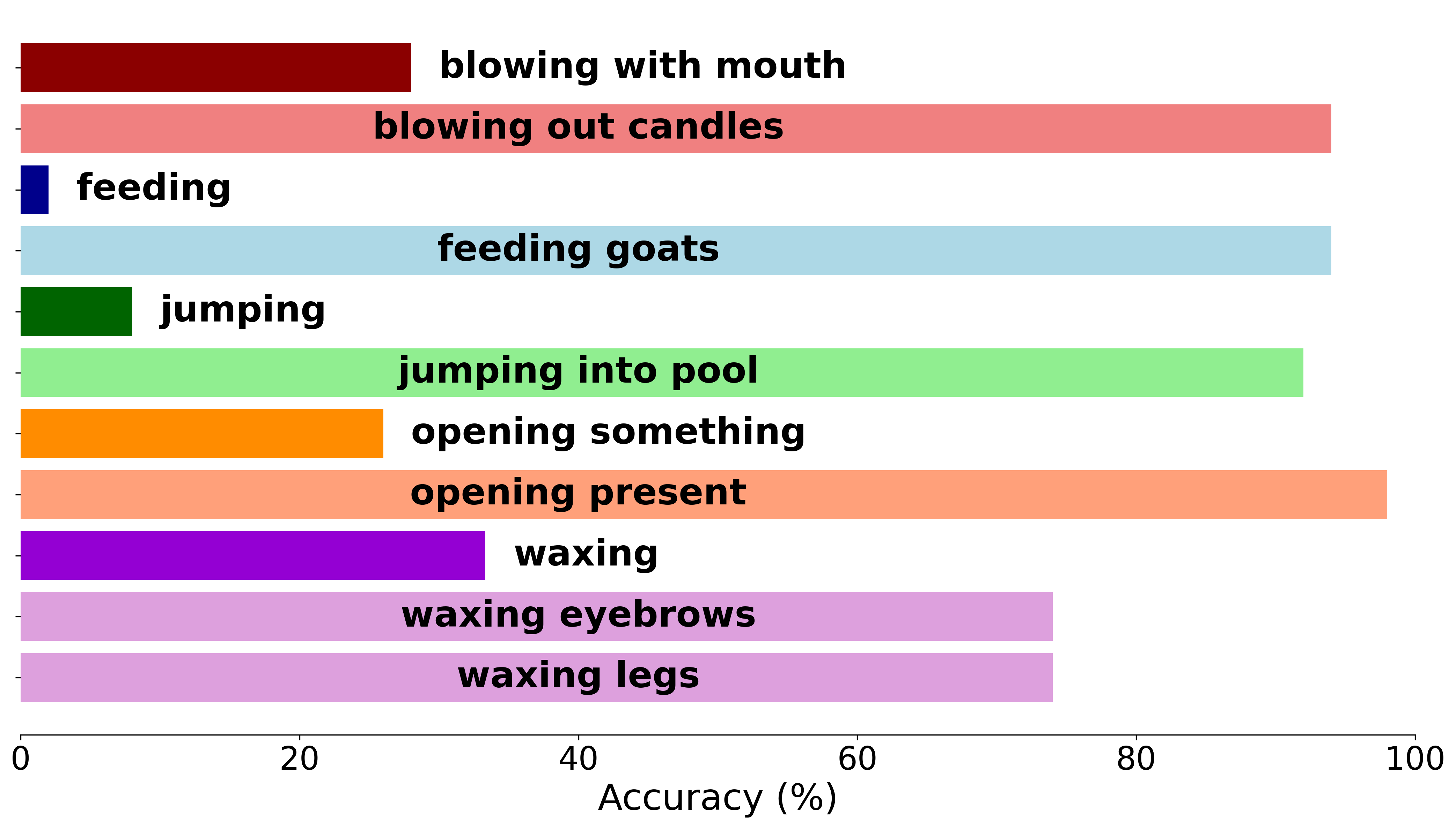}
\vspace{-20pt}
\caption{\textbf{\textit{Example cases of K400-TA where unknown fine performance is higher:}} The bars with a darker color denote coarse classes while their corresponding fine class is shown in a lighter color. The accuracy of coarse classes is noticably lower than fine counterparts. Results are shown for EZ-CLIP model.
}
 \label{fig:k400_classwise_set1_ezclip}
\end{figure}

\begin{figure}[!t]
\centering
\includegraphics[width=1\linewidth]{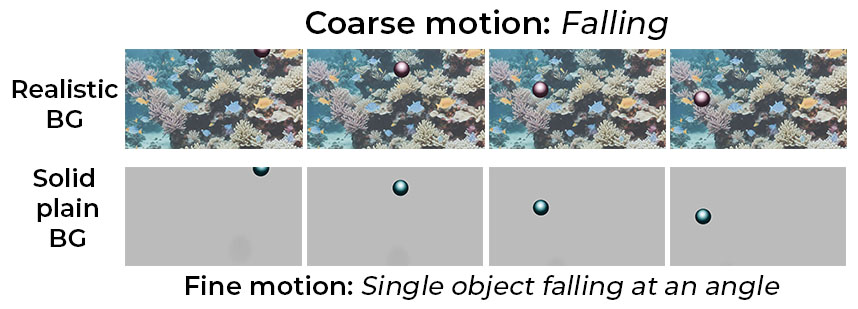}
\vspace{-20pt}
\caption{\textbf{\textit{Preview of Syn-TA realistic vs. solid plain background:}} The absence of complex textures in the background enhances the model’s ability to comprehend object motion. This suggests that even for relatively simple motions, models are often challenged by textured backgrounds.}
 \label{fig:dataset_preview_Syn-TA_realistic_vs_white}
\end{figure}

\label{supp_disentanglement}
We tested our approach on EZ-CLIP \cite{ezclip}, chosen for its efficiency due to fewer learnable parameters, achieving competitive results with reduced time and computation. We modified EZ-CLIP's video encoder by adding two branches in the final layers, duplicating the transformer blocks in the last two layers to separate feature learning. The earlier layers capture low and mid-level features, while one branch of the final layers specializes in high-level (coarse) concepts (e.g., \textit{``Pushing''}) and the other focuses on fine-grained context (e.g., \textit{``Pushing something from left to right''}). As shown in \cref{fig:archi_improvement}, the fine feature is combined with the coarse feature at every step. The class token (CLS) from each branch flows through a shared normalization layer and distinct projection layers (fine one is trainable), generating two embeddings. We also disabled passing EZ-CLIP's temporal prompt in the final two layers. This setup allows the coarse branch to focus on high-level features while the fine branch selectively contributes necessary detail, improving overall coarse accuracy and maintaining focused feature disentanglement in each branch. As shown in \cref{tab:score_improvement_base_vs_ours_detailed}, our approach improves coarse motion accuracy on temporal datasets such as Syn-TA and SSv2-TA.

\clearpage

\section{Detailed list of classes}
\label{supp_classes}
The full list of coarse classes and fine classes for each dataset is provided in this section. The split files for each of the dataset is also provided in our \href{https://github.com/raiyaan-abdullah/Motion-Transfer}{GitHub}.

\subsection{Syn-TA}
In our newly proposed dataset `Syn-TA', we generated videos of plain 3D objects, such as cubes, spheres, cylinders, etc, performing various motions using the 3D modeling software Blender \cite{blender}. Each video features a realistic background depicting either outdoor scenes (e.g., desert, forest, sunset) or indoor settings (e.g., coffee shop, kitchen, library), where 3D objects perform various motions. The videos are rendered at 24 frames per second (FPS) with a resolution of 1920x1080 pixels. The camera view is either from the front or the top. \\
The dataset includes 20 coarse motions, which are further subdivided into 100 fine motions. Each video contains, on average, 105 frames with a standard deviation of approximately 46. The fine motions are split into two subsets: $S1$ (53 classes) and $S2$ (47 classes). Each set has its own collection of object shapes and backgrounds, creating a more challenging setting for evaluating model performance on novel scenarios. Each fine motion is labeled with its corresponding coarse class and additional contextual details. Both subsets include at least two fine classes under each coarse motion. If a coarse class contains an even number of fine classes, they are evenly distributed between $S1$ and $S2$. For coarse classes with an odd number of fine motions, $S1$ is assigned one extra class. For example, for the coarse class \textit{``departing''}, the fine class \textit{```multiple objects departing simultaneously''} is in $S1$ while \textit{``multiple objects departing one by one''} is in $S2$. \\
This carefully constructed dataset is intended to serve as a diagnostic benchmark for evaluating how models adapt to detecting high-level actions under varying contexts. The Blender Python API code used to generate the videos is available  in our \href{https://github.com/raiyaan-abdullah/Motion-Transfer}{GitHub}, and samples of each class are provided. The videos are stored in standard .mp4 format and follow the structure of Kinetics400 \cite{kinetics}, ensuring compatibility with existing dataloader implementations. \\

\vfill\null

\begin{table}[H]
\centering
\caption{List of coarse classes for Syn-TA}
\label{tab:3d_object_motions_head_classes} 
\begin{tabular}{cl}
\hline
\textbf{Coarse ID} & \textbf{Coarse motion} \\
\hline
0 & appearing \\ \hline
1 & arriving \\ \hline
2 & bouncing \\ \hline
3 & changing color \\ \hline
4 & changing shape \\ \hline
5 & cloning \\ \hline
6 & departing \\ \hline
7 & disappearing \\ \hline
8 & expanding \\ \hline
9 & falling \\ \hline
10 & following \\ \hline
11 & merging \\ \hline
12 & moving in a path \\ \hline
13 & not colliding \\ \hline
14 & occlusion \\ \hline
15 & orbiting \\ \hline
16 & rising \\ \hline
17 & shooting projectile \\ \hline
18 & shrinking \\ \hline
19 & teleporting \\ \hline
\end{tabular}
\end{table}

\clearpage

\onecolumn
\begin{longtable}{clcp{7cm}}

\caption{List of the fine classes in Syn-TA - Set 1}
\label{tab:3d_objects_set1_full} \\

\hline
\textbf{Coarse ID} & \textbf{Coarse motion} & \textbf{Fine ID} & \textbf{Coarse motion + fine motion} \\ \hline
\endfirsthead

\hline
\textbf{Coarse ID} & \textbf{Coarse motion} & \textbf{Fine ID} & \textbf{Coarse motion + fine motion} \\ \hline
\endhead

\hline \multicolumn{4}{r}{{Continued on next page}} \\ \hline
\endfoot

\hline
\endlastfoot

\hline
\multirow{2}{*}{0}  & \multirow{2}{*}{appearing}           & 0  & Single object appearing by increasing opacity                                                                                             \\ \cline{3-4}
                    &                                      & 1  & Multiple objects all appearing simultaneously by increasing opacity                                                                       \\ \hline
\multirow{3}{*}{1}  & \multirow{3}{*}{arriving}            & 2  & Single object arriving                                                                                                                    \\ \cline{3-4}
                    &                                      & 3  & Multiple objects arriving simultaneously                                                                                                  \\ \cline{3-4}
                    &                                      & 4  & Single object arriving beside another single object                                                                                       \\ \hline
\multirow{4}{*}{2}  & \multirow{4}{*}{bouncing}            & 5  & Single ball bouncing on a plain surface eventually stops                                                                                  \\ \cline{3-4}
                    &                                      & 6  & Multiple balls bouncing on plain surface at very different heights eventually stop                                                        \\ \cline{3-4}
                    &                                      & 7  & Single object bouncing and going forward                                                                                                   \\ \cline{3-4}
                    &                                      & 8  & Single object imitating the motion of bouncing and going down stairs                                                                        \\ \hline
\multirow{3}{*}{3}  & \multirow{3}{*}{changing color}      & 9  & Single object keeps changing colors                                                                                                       \\ \cline{3-4}
                    &                                      & 10 & Two objects changing colors by switching with each other                                                                                  \\ \cline{3-4}
                    &                                      & 11 & Single object changing from bright to dark color                                                                                          \\ \hline
\multirow{4}{*}{4}  & \multirow{3}{*}{changing shape}      & 12 & Single object of other shape changes into pyramid                                                                                          \\ \cline{3-4}
                    &                                      & 13 & Single object of other shape changes into cube                                                                                            \\ \cline{3-4}
                    &                                      & 14 & Two objects changing shapes by switching with each other                                                                                  \\ \hline
\multirow{3}{*}{5}  & \multirow{3}{*}{cloning}             & 15 & Single object cloning into two identical objects both going in same direction                                                             \\ \cline{3-4}
                    &                                      & 16 & Single object cloning into multiple identical objects one by one                                                                          \\ \cline{3-4}
                    &                                      & 17 & Single object cloning into one identical object and another object with different colour                                                  \\ \hline
\multirow{3}{*}{6}  & \multirow{3}{*}{departing}           & 18 & Single object departing                                                                                                                   \\ \cline{3-4}
                    &                                      & 19 & Multiple objects departing simultaneously                                                                                                 \\ \cline{3-4}
                    &                                      & 20 & Single object departing beside another single object                                                                                      \\ \hline
\multirow{2}{*}{7}  & \multirow{2}{*}{disappearing}        & 21 & Single object disappearing by decreasing opacity                                                                                          \\ \cline{3-4}
                    &                                      & 22 & Multiple objects all disappearing simultaneously by decreasing opacity                                                                    \\ \hline
\multirow{2}{*}{8}  & \multirow{2}{*}{expanding}           & 23 & Small single object expanding into medium object                                                                                          \\ \cline{3-4}
                    &                                      & 24 & Two different small objects expanding into similar larger size                                                                               \\ \hline
\multirow{2}{*}{9}  & \multirow{2}{*}{falling}             & 25 & Single object falling straight down without changing direction                                                                            \\ \cline{3-4}
                    &                                      & 26 & Single object falling down in a zigzag pattern                                                                                            \\ \hline
\multirow{3}{*}{10} & \multirow{3}{*}{following}           & 27 & Single object following another object all the time while moving in a straight line                                                       \\ \cline{3-4}
                    &                                      & 28 & Single object following another object for some time while moving in straight line then follows another path                              \\ \cline{3-4}
                    &                                      & 29 & Multiple objects following one object all the time in a straight line                                                                     \\ \hline
\multirow{2}{*}{11} & \multirow{2}{*}{merging}             & 30 & Two objects merging into one similar sized object while moving towards each other                                                         \\ 
                    &                                      & 31 & Multiple objects merging into one big object                                                                                              \\ \hline
\multirow{3}{*}{12} & \multirow{3}{*}{moving in a path}    & 32 & Single object moving slowly from one position to another and stopping                                                                     \\ \cline{3-4}
                    &                                      & 33 & Single object moving slowly from one position to another and coming back to original position                                             \\ \cline{3-4}
                    &                                      & 34 & Single object moving from left to right                                                                                                   \\ \hline
\multirow{3}{*}{13} & \multirow{3}{*}{not colliding}       & 35 & Two objects moving towards each other but not colliding                                                                                   \\ \cline{3-4}
                    &                                      & 36 & Single object moving towards single stationary object but not colliding with it                                                           \\ \cline{3-4}
                    &                                      & 37 & Multiple objects moving towards each other but not colliding among themselves                                                             \\ \hline
\multirow{2}{*}{14} & \multirow{2}{*}{occlusion}           & 38 & Small object coming in front of big object barely occluding it                                                                             \\ \cline{3-4}
                    &                                      & 39 & Object which is occluding another object switches places and goes behind it                                                               \\ \hline
\multirow{3}{*}{15} & \multirow{3}{*}{orbiting}            & 40 & Single object orbiting another object in the center in clockwise direction on horizontal plane                                            \\ \cline{3-4}
                    &                                      & 41 & Single object orbiting another object in the center in clockwise direction first and then counter clockwise direction on horizontal plane \\ \cline{3-4}
                    &                                      & 42 & Single object orbiting another object in the center in clockwise direction on vertical plane                                              \\ \hline
\multirow{2}{*}{16} & \multirow{2}{*}{rising}              & 43 & Single object rising straight up without changing direction                                                                               \\ \cline{3-4}
                    &                                      & 44 & Single object rising in a zigzag pattern                                                                                                  \\ \hline
\multirow{3}{*}{17} & \multirow{3}{*}{shooting projectile} & 45 & Shooting projectile from one corner to opposite corner                                                                                    \\ \cline{3-4}
                    &                                      & 46 & Shooting projectile from right corner to left corner                                                                                      \\ \cline{3-4}
                    &                                      & 47 & Shooting two projectiles from same position to opposite corner                                                                            \\ \hline
\multirow{2}{*}{18} & \multirow{2}{*}{shrinking}           & 48 & Large single object shrinking into medium sized object                                                                                    \\ \cline{3-4}
                    &                                      & 49 & Two different large objects shrinking into same smaller size                                                                              \\ \hline
\multirow{3}{*}{19} & \multirow{3}{*}{teleporting}         & 50 & Single object teleporting from one corner to another                                                                                      \\ \cline{3-4}
                    &                                      & 51 & Single object teleporting from one corner to another while another object is in the center                                                \\ \cline{3-4}
                    &                                      & 52 & Two objects in opposite corners switch places by teleporting                                                                              \\ \hline
\end{longtable}

\begin{longtable}{clcp{7cm}}

\caption{List of the classes in Syn-TA - Set 2}
\label{tab:3d_objects_set2_full} \\

\hline
\textbf{Coarse ID} & \textbf{Coarse motion} & \textbf{Fine ID} & \textbf{Coarse motion + fine motion} \\ \hline
\endfirsthead

\hline
\textbf{Coarse ID} & \textbf{Coarse motion} & \textbf{Fine ID} & \textbf{Coarse motion + fine motion} \\ \hline
\endhead

\hline \multicolumn{4}{r}{{Continued on next page}} \\ \hline
\endfoot

\hline
\endlastfoot

\hline
\multirow{2}{*}{0}  & \multirow{2}{*}{appearing}           & 0  & Multiple objects all appearing one by one by increasing opacity                                                                                             \\ \cline{3-4}
                    &                                      & 1  & Two objects appearing among two other objects                                                                       \\ \hline
\multirow[t]{3}{*}{1}  & \multirow[t]{3}{*}{arriving}            & 2  & Multiple objects arriving one by one                                                                                                                    \\ \cline{3-4}
                    &                                      & 3  & Single object arriving among many objects                                                                                                  \\ \cline{3-4}
                    &                                      & 4  & Single object arriving beside another single object, pausing then going away                                                                                       \\ \hline
\multirow{4}{*}{2}  & \multirow{4}{*}{bouncing}            & 5  & Multiple objects bouncing on plain surface at almost similar heights eventually stop                                                                                  \\ \cline{3-4}
                    &                                      & 6  & Single object imitating the motion of bouncing against a wall                                                        \\ \cline{3-4}
                    &                                      & 7  & Single object bouncing on a plain surface perpetually                                                                                                    \\ \cline{3-4}
                    &                                      & 8  & Single object imitating the motion of bouncing and going up stairs                                                                        \\ \hline
\multirow{3}{*}{3}  & \multirow{3}{*}{changing color}      & 9  & Single object changing from dark to bright color                                                                                                       \\ \cline{3-4}
                    &                                      & 10 & Single object changing color then returning to previous one                                                                                  \\ \hline
\multirow{3}{*}{4}  & \multirow{2}{*}{changing shape}      & 11 & Single object of other shape changes into sphere                                                                                          \\ \cline{3-4}
                    &                                      & 12 & Single object of other shape changes into cylinder                                                                                            \\ 
                    &                                      & 13 & Single object changing into another shape then turning back to original shape                                                                                \\ \hline
\multirow{2}{*}{5}  & \multirow{2}{*}{cloning}             & 14 & Single object cloning into two identical objects both going in different directions                                                             \\ \cline{3-4}
                    &                                      & 15 & Single object cloning into multiple identical objects at once                                                                          \\ \hline
\multirow{3}{*}{6}  & \multirow{3}{*}{departing}           & 16 & Multiple objects departing one by one                                                                                                                   \\ \cline{3-4}
                    &                                      & 17 & Single object departing among many objects                                                                                                 \\ \cline{3-4}
                    &                                      & 18 & Single object departing beside another single object, then coming back shortly                                                                                      \\ \hline
\multirow{2}{*}{7}  & \multirow{2}{*}{disappearing}        & 19 & Multiple objects all disappearing one by one by decreasing opacity                                                                                          \\ \cline{3-4}
                    &                                      & 20 & Multiple objects some disappearing and some remaining                                                                    \\ \hline
\multirow{2}{*}{8}  & \multirow{2}{*}{expanding}           & 21 & Small single object expanding into large object                                                                                          \\ \cline{3-4}
                    &                                      & 22 & Two same small objects expanding into different bigger sizes                                                                               \\ \hline
\multirow{2}{*}{9}  & \multirow{2}{*}{falling}             & 23 & Single object falling at an angle                                                                            \\ \cline{3-4}
                    &                                      & 24 & Single object falling at a curved path                                                                                            \\ \hline
\multirow{2}{*}{10} & \multirow{2}{*}{following}           & 25 & Single object following another object all the time in a complex path                                                       \\ \cline{3-4}
                    &                                      & 26 & Single object not following another object at first but later starts following                                                                     \\ \hline
\multirow{2}{*}{11} & \multirow{2}{*}{merging}             & 27 & Faster object merging with slower object into one similar sized object while moving in same direction                                                         \\ \cline{3-4}
                    &                                      & 28 & Multiple objects merging into multiple big objects                                                                                              \\ \hline
\multirow{3}{*}{12} & \multirow{3}{*}{moving in a path}    & 29 & Single object moving quickly from one position to another and stopping                                                                     \\ \cline{3-4}
                    &                                      & 30 & Two objects moving and switching their positions                                             \\ \cline{3-4}
                    &                                      & 31 & Single object moving from right to left                                                                                                   \\ \hline
\multirow{2}{*}{13} & \multirow{2}{*}{not colliding}       & 32 & Two objects moving in opposite directions in circular path but not colliding later                                                                                   \\ \cline{3-4}
                    &                                      & 33 & One object moving towards multiple stationary objects but not colliding                                                           \\ \hline
\multirow{2}{*}{14} & \multirow{2}{*}{occlusion}           & 34 & Big object coming in front of small object mostly or fully occluding it                                                                            \\ \cline{3-4}
                    &                                      & 35 & Two objects partially occluding two other objects switch positions with the latter two and go behind them                                                               \\ \hline
\multirow{3}{*}{15} & \multirow{3}{*}{orbiting}            & 36 & Single object orbiting another object in the center in counter clockwise direction on horizontal plane                                            \\ \cline{3-4}
                    &                                      & 37 & Single object orbiting another object in clockwise direction then second object orbiting first object in clockwise direction on horizontal plane \\ \cline{3-4}
                    &                                      & 38 & Single object orbiting another object in the center in counter clockwise direction on vertical plane                                              \\ \hline
\multirow{2}{*}{16} & \multirow{2}{*}{rising}              & 39 & Single object rising at an angle                                                                               \\ \cline{3-4}
                    &                                      & 40 & Single object rising at a curved path                                                                                                  \\ \hline
\multirow{2}{*}{17} & \multirow{2}{*}{shooting projectile} & 41 & Shooting projectile from left corner to right corner                                                                                    \\ \cline{3-4}
                    &                                      & 42 & Shooting two projectiles from opposite corners switching their positions                                                                            \\ \hline
\multirow{2}{*}{18} & \multirow{2}{*}{shrinking}           & 43 & Large single object shrinking into very small object                                                                                    \\ \cline{3-4}
                    &                                      & 44 & Two different large objects shrinking into different smaller sizes                                                                              \\ \hline
\multirow{2}{*}{19} & \multirow{2}{*}{teleporting}         & 45 & Four objects in four corners switch places by teleporting                                                                                      \\ \cline{3-4}
                    &                                      & 46 & Two objects together in one corner then one teleports to other corner                                                                                      \\ \hline
\end{longtable}

\subsection{Kinetics400 - Transferable Activity}

\begin{longtable}{cl}
\caption{List of coarse classes for K400-TA}
\label{tab:k400_head_classes} \\

\hline
\textbf{Coarse ID} & \textbf{Coarse motion} \\
\hline
\endfirsthead

\hline
\textbf{Coarse ID} & \textbf{Coarse motion} \\
\hline
\endhead

\hline
\endfoot

\hline
\endlastfoot

0           & blowing with mouth       \\ \hline
1           & building a structure     \\ \hline
2           & catching or throwing other objects than ball  \\ \hline
3           & cleaning                 \\ \hline
4           & climbing                 \\ \hline
5           & cooking                  \\ \hline
6           & dancing                  \\ \hline
7           & drinking                 \\ \hline
8           & eating                   \\ \hline
9           & feeding                  \\ \hline
10          & fishing                  \\ \hline
11          & folding                  \\ \hline
12          & gardening                \\ \hline
13          & grooming animal          \\ \hline
14          & hair care                \\ \hline
15          & juggling                 \\ \hline
16          & jumping                  \\ \hline
17          & kicking                  \\ \hline
18          & maintaining vehicle      \\ \hline
19          & massaging                \\ \hline
20          & music without instrument \\ \hline
21          & opening something        \\ \hline
22          & petting animal           \\ \hline
23          & playing by hitting ball with something        \\ \hline
24          & playing instrument       \\ \hline
25          & playing with ball in hand                     \\ \hline
26          & preparing fruit or vegetable                  \\ \hline
27          & punching                 \\ \hline
28          & pushing                  \\ \hline
29          & reading                  \\ \hline
30          & riding animal            \\ \hline
31          & riding on something over water                 \\ \hline
32          & riding vehicle           \\ \hline
33          & shaving                  \\ \hline
34          & skiing                   \\ \hline
35          & smoking                  \\ \hline
36          & swimming                 \\ \hline
37          & toddler interaction      \\ \hline
38          & tying                    \\ \hline
39          & washing                  \\ \hline
40          & waxing                   \\ \hline
\end{longtable}

\begin{longtable}{clcl}
\caption{List of the fine classes in K400-TA Set 1}
\label{tab:k400_set1_full} \\

\hline
\textbf{Coarse ID} & \textbf{Coarse motion} & \textbf{Fine ID} & \textbf{Coarse motion + fine motion} \\ \hline
\endfirsthead

\hline
\textbf{Coarse ID} & \textbf{Coarse motion} & \textbf{Fine ID} & \textbf{Coarse motion + fine motion} \\ \hline
\endhead

\hline \multicolumn{4}{r}{{Continued on next page}} \\ \hline
\endfoot

\hline
\endlastfoot

0                    & blowing with mouth                                            & 0   & balloon blowing                     \\ \hline
1                    & building a structure                                          & 1   & building cabinet                    \\ \hline
\multirow{2}{*}{2}   & catching or throwing                                          & 2   & catching or throwing frisbee        \\ \cline{3-4}
{}                   & other objects than ball                                       & 3   & throwing axe                        \\ \hline
\multirow{3}{*}{3}   & \multirow{3}{*}{cleaning}                                     & 4   & cleaning floor                      \\ \cline{3-4}
                     &                                                               & 5   & cleaning gutters                    \\ \cline{3-4}
                     &                                                               & 6   & cleaning pool                       \\ \hline
\multirow{2}{*}{4}   & \multirow{2}{*}{climbing}                                     & 7   & climbing a rope                     \\ \cline{3-4}
                     &                                                               & 8   & climbing ladder                     \\ \hline
\multirow{4}{*}{5}   & \multirow{4}{*}{cooking}                                      & 9   & baking cookies                      \\ \cline{3-4}
                     &                                                               & 10  & barbequing                          \\ \cline{3-4}
                     &                                                               & 11  & cooking chicken                     \\ \cline{3-4}
                     &                                                               & 12  & cooking egg                         \\ \hline
\multirow{8}{*}{6}   & \multirow{8}{*}{dancing}                                      & 13  & belly dancing                       \\ \cline{3-4}
                     &                                                               & 14  & breakdancing                        \\ \cline{3-4}
                     &                                                               & 15  & country line dancing                \\ \cline{3-4}
                     &                                                               & 16  & dancing ballet                      \\ \cline{3-4}
                     &                                                               & 17  & dancing charleston                  \\ \cline{3-4}
                     &                                                               & 18  & dancing gangnam style               \\ \cline{3-4}
                     &                                                               & 19  & dancing macarena                    \\ \cline{3-4}
                     &                                                               & 20  & jumpstyle dancing                   \\ \hline
\multirow{2}{*}{7}   & \multirow{2}{*}{drinking}                                     & 21  & drinking beer                       \\ \cline{3-4}
                     &                                                               & 22  & drinking shots                      \\ \hline
\multirow[t]{6}{*}{8}   & \multirow[t]{6}{*}{eating}                                       & 23  & dining                              \\ \cline{3-4}
                     &                                                               & 24  & eating burger                       \\ \cline{3-4}
                     &                                                               & 25  & eating cake                         \\ \cline{3-4}
                     &                                                               & 26  & eating carrots                      \\ \cline{3-4}
                     &                                                               & 27  & eating chips                        \\ \cline{3-4}
                     &                                                               & 28  & eating doughnuts                    \\ \hline
\multirow{2}{*}{9}   & \multirow{2}{*}{feeding}                                      & 29  & feeding birds                       \\ \cline{3-4}
                     &                                                               & 30  & feeding fish                        \\ \hline
10                   & fishing                                                       & 31  & catching fish                       \\ \hline
\multirow{2}{*}{11}  & \multirow{2}{*}{folding}                                      & 32  & folding clothes                     \\ \cline{3-4}
                     &                                                               & 33  & folding napkins                     \\ \hline
\multirow{3}{*}{12}  & \multirow{3}{*}{gardening}                                    & 34  & arranging flowers                   \\ \cline{3-4}
                     &                                                               & 35  & planting trees                      \\ \cline{3-4}
                     &                                                               & 36  & stomping grapes                     \\ \hline
13                   & grooming animal                                               & 37  & grooming dog                        \\ \hline
\multirow{4}{*}{14}  & \multirow{4}{*}{hair care}                                    & 38  & braiding hair                       \\ \cline{3-4}
                     &                                                               & 39  & brushing hair                       \\ \cline{3-4}
                     &                                                               & 40  & curling hair                        \\ \cline{3-4}
                     &                                                               & 41  & dying hair                          \\ \hline
\multirow{2}{*}{15}  & \multirow{2}{*}{juggling}                                     & 42  & contact juggling                    \\ \cline{3-4}
                     &                                                               & 43  & juggling balls                      \\ \hline
16                   & jumping                                                       & 44  & high jump                           \\ \hline
\multirow{4}{*}{17}  & \multirow{4}{*}{kicking}                                      & 45  & drop kicking                        \\ \cline{3-4}
                     &                                                               & 46  & high kick                           \\ \cline{3-4}
                     &                                                               & 47  & kicking field goal                  \\ \cline{3-4}
                     &                                                               & 48  & kicking soccer ball                 \\ \hline
\multirow{2}{*}{18}  & \multirow{2}{*}{maintaining vehicle}                          & 49  & changing oil                        \\ \cline{3-4}
                     &                                                               & 50  & changing wheel                      \\ \hline
\multirow{2}{*}{19}  & \multirow{2}{*}{massaging}                                    & 51  & massaging back                      \\ \cline{3-4}
                     &                                                               & 52  & massaging feet                      \\ \hline
\multirow{2}{*}{20}  & \multirow{2}{*}{music without instrument}                     & 53  & air drumming                        \\ \cline{3-4}
                     &                                                               & 54  & beatboxing                          \\ \hline
21                   & opening something                                             & 55  & opening bottle                      \\ \hline
22                   & petting animal                                                & 56  & petting animal (not cat)            \\ \hline
\multirow{3}{*}{23}  & playing by hitting                                            & 57  & golf chipping                       \\ \cline{3-4}
                     & ball with something                                           & 58  & golf driving                        \\ \cline{3-4}
                     & {}                                                            & 59  & hitting baseball                    \\ \hline
\multirow{13}{*}{24} & \multirow{13}{*}{playing instrument}                          & 60  & busking                             \\ \cline{3-4}
                     &                                                               & 61  & drumming fingers                    \\ \cline{3-4}
                     &                                                               & 62  & playing accordion                   \\ \cline{3-4}
                     &                                                               & 63  & playing bagpipes                    \\ \cline{3-4}
                     &                                                               & 64  & playing bass guitar                 \\ \cline{3-4}
                     &                                                               & 65  & playing cello                       \\ \cline{3-4}
                     &                                                               & 66  & playing clarinet                    \\ \cline{3-4}
                     &                                                               & 67  & playing cymbals                     \\ \cline{3-4}
                     &                                                               & 68  & playing didgeridoo                  \\ \cline{3-4}
                     &                                                               & 69  & playing drums                       \\ \cline{3-4}
                     &                                                               & 70  & playing flute                       \\ \cline{3-4}
                     &                                                               & 71  & playing guitar                      \\ \cline{3-4}
                     &                                                               & 72  & playing harmonica                   \\ \hline
\multirow[t]{6}{*}{25}  & \multirow[t]{6}{*}{playing with ball in hand}                    & 73  & catching or throwing baseball       \\ \cline{3-4}
                     &                                                               & 74  & dodgeball                           \\ \cline{3-4}
                     &                                                               & 75  & dribbling basketball                \\ \cline{3-4}
                     &                                                               & 76  & dunking basketball                  \\ \cline{3-4}
                     &                                                               & 77  & passing American football (in game) \\ \cline{3-4}
                     &                                                               & 78  & throwing ball                       \\ \hline
\multirow{2}{*}{26}  & \multirow{2}{*}{preparing fruit or vegetable}                 & 79  & cutting pineapple                   \\ \cline{3-4}
                     &                                                               & 80  & peeling apples                      \\ \hline
27                   & punching                                                      & 81  & punching bag                        \\ \hline
\multirow{2}{*}{28}  & \multirow{2}{*}{pushing}                                      & 82  & pushing car                         \\ \cline{3-4}
                     &                                                               & 83  & pushing cart                        \\ \hline
29                   & reading                                                       & 84  & reading book                        \\ \hline
\multirow{2}{*}{30}  & \multirow{2}{*}{riding animal}                                & 85  & riding camel                        \\ \cline{3-4}
                     &                                                               & 86  & riding elephant                     \\ \hline
\multirow{3}{*}{31}  & \multirow{3}{*}{riding on something over water}               & 87  & crossing river                      \\ \cline{3-4}
                     &                                                               & 88  & sailing                             \\ \cline{3-4}
                     &                                                               & 89  & surfing water                       \\ \hline
\multirow{5}{*}{32}  & \multirow{5}{*}{riding vehicle}                               & 90  & biking through snow                 \\ \cline{3-4}
                     &                                                               & 91  & driving car                         \\ \cline{3-4}
                     &                                                               & 92  & motorcycling                        \\ \cline{3-4}
                     &                                                               & 93  & riding scooter                      \\ \cline{3-4}
                     &                                                               & 94  & snowmobiling                        \\ \hline
33                   & shaving                                                       & 95  & shaving head                        \\ \hline
\multirow{3}{*}{34}  & \multirow{3}{*}{skiing}                                       & 96  & ski jumping                         \\ \cline{3-4}
                     &                                                               & 97  & skiing (not slalom or crosscountry) \\ \cline{3-4}
                     &                                                               & 98  & water skiing                        \\ \hline
35                   & smoking                                                       & 99  & smoking cigarette or short object   \\ \hline
\multirow{3}{*}{36}  & \multirow{3}{*}{swimming}                                     & 100 & scuba diving                        \\ \cline{3-4}
                     &                                                               & 101 & swimming backstroke                 \\ \cline{3-4}
                     &                                                               & 102 & swimming breast stroke              \\ \hline
\multirow{2}{*}{37}  & \multirow{2}{*}{toddler interaction}                          & 103 & baby waking up                      \\ \cline{3-4}
                     &                                                               & 104 & carrying baby                       \\ \hline
\multirow{2}{*}{38}  & \multirow{2}{*}{tying}                                        & 105 & tying bow tie                       \\ \cline{3-4}
                     &                                                               & 106 & tying knot (not on a tie)           \\ \hline
\multirow{2}{*}{39}  & \multirow{2}{*}{washing}                                      & 107 & washing dishes                      \\ \cline{3-4}
                     &                                                               & 108 & washing feet                        \\ \hline
\multirow{2}{*}{40}  & \multirow{2}{*}{waxing}                                       & 109 & waxing back                         \\ \cline{3-4}
                     &                                                               & 110 & waxing chest                        \\ \hline

\end{longtable}

\begin{longtable}{clcp{5cm}}
\caption{List of the fine classes in K400-TA Set 2}
\label{tab:k400_set2_full} \\

\hline
\textbf{Coarse ID} & \textbf{Coarse motion} & \textbf{Fine ID} & \textbf{Coarse motion + fine motion} \\ \hline
\endfirsthead

\hline
\textbf{Coarse ID} & \textbf{Coarse motion} & \textbf{Fine ID} & \textbf{Coarse motion + fine motion} \\ \hline
\endhead

\hline \multicolumn{4}{r}{{Continued on next page}} \\ \hline
\endfoot

\hline
\endlastfoot

0                    & blowing with mouth                                      & 0  & blowing out candles                     \\ \hline
1                    & building a structure                                    & 1  & building shed                           \\ \hline
2                    & catching or throwing                                    & 2  & throwing discus                         \\
{}                   & other objects than ball                                 & {} & {}                                      \\ \hline
\multirow{3}{*}{3}   & \multirow{3}{*}{cleaning}                               & 3  & cleaning shoes                          \\ \cline{3-4}
                     &                                                         & 4  & cleaning toilet                         \\ \cline{3-4}
                     &                                                         & 5  & cleaning windows                        \\ \hline
\multirow{2}{*}{4}   & \multirow{2}{*}{climbing}                               & 6  & climbing tree                           \\ \cline{3-4}
                     &                                                         & 7  & ice climbing                            \\ \hline
\multirow[t]{4}{*}{5}   & \multirow[t]{4}{*}{cooking}                                & 8  & cooking on campfire                     \\ \cline{3-4}
                     &                                                         & 9  & cooking sausages                        \\ \cline{3-4}
                     &                                                         & 10 & flipping pancake                        \\ \cline{3-4}
                     &                                                         & 11 & frying vegetables                       \\ \hline
\multirow{7}{*}{6}   & \multirow{7}{*}{dancing}                                & 12 & krumping                                \\ \cline{3-4}
                     &                                                         & 13 & robot dancing                           \\ \cline{3-4}
                     &                                                         & 14 & salsa dancing                           \\ \cline{3-4}
                     &                                                         & 15 & swing dancing                           \\ \cline{3-4}
                     &                                                         & 16 & tango dancing                           \\ \cline{3-4}
                     &                                                         & 17 & tap dancing                             \\ \cline{3-4}
                     &                                                         & 18 & zumba                                   \\ \hline
\multirow{2}{*}{7}   & \multirow{2}{*}{drinking}                               & 19 & drinking water or soft drinks           \\ \cline{3-4}
                     &                                                         & 20 & tasting beer                            \\ \hline
\multirow{5}{*}{8}   & \multirow{5}{*}{eating}                                 & 21 & eating hotdog                           \\ \cline{3-4}
                     &                                                         & 22 & eating ice cream                        \\ \cline{3-4}
                     &                                                         & 23 & eating spaghetti                        \\ \cline{3-4}
                     &                                                         & 24 & eating watermelon                       \\ \cline{3-4}
                     &                                                         & 25 & tasting food                            \\ \hline
9                    & feeding                                                 & 26 & feeding goats                           \\ \hline
10                   & fishing                                                 & 27 & ice fishing                             \\ \hline
11                   & folding                                                 & 28 & folding paper                           \\ \hline
\multirow{2}{*}{12}  & \multirow{2}{*}{gardening}                              & 29 & trimming trees                          \\ \cline{3-4}
                     &                                                         & 30 & watering plants                         \\ \hline
13                   & grooming animal                                         & 31 & grooming horse                          \\ \hline
\multirow{3}{*}{14}  & \multirow{3}{*}{hair care}                              & 32 & fixing hair                             \\ \cline{3-4}
                     &                                                         & 33 & getting a haircut                       \\ \cline{3-4}
                     &                                                         & 34 & trimming or shaving beard               \\ \hline
\multirow{2}{*}{15}  & \multirow{2}{*}{juggling}                               & 35 & juggling fire                           \\ \cline{3-4}
                     &                                                         & 36 & juggling soccer ball                    \\ \hline
16                   & jumping                                                 & 37 & jumping into pool                       \\ \hline
\multirow{3}{*}{17}  & \multirow{3}{*}{kicking}                                & 38 & playing kickball                        \\ \cline{3-4}
                     &                                                         & 39 & shooting goal (soccer)                  \\ \cline{3-4}
                     &                                                         & 40 & side kick                               \\ \hline
\multirow{2}{*}{18}  & \multirow{2}{*}{maintaining vehicle}                    & 41 & checking tires                          \\ \cline{3-4}
                     &                                                         & 42 & pumping gas                             \\ \hline
\multirow{2}{*}{19}  & \multirow{2}{*}{massaging}                              & 43 & massaging legs                          \\ \cline{3-4}
                     &                                                         & 44 & massaging person's head                 \\ \hline
20                   & music without instrument                                & 45 & singing                                 \\ \hline
21                   & opening something                                       & 46 & opening present                         \\ \hline
22                   & petting animal                                          & 47 & petting cat                             \\ \hline
\multirow{2}{*}{23}  & playing by hitting                                      & 48 & golf putting                            \\ \cline{3-4}
                     & ball with something                                     & 49 & playing cricket                         \\ \hline
\multirow{13}{*}{24} & \multirow{13}{*}{playing instrument}                    & 50 & playing harp                            \\ \cline{3-4}
                     &                                                         & 51 & playing keyboard                        \\ \cline{3-4}
                     &                                                         & 52 & playing organ                           \\ \cline{3-4}
                     &                                                         & 53 & playing piano                           \\ \cline{3-4}
                     &                                                         & 54 & playing recorder                        \\ \cline{3-4}
                     &                                                         & 55 & playing saxophone                       \\ \cline{3-4}
                     &                                                         & 56 & playing trombone                        \\ \cline{3-4}
                     &                                                         & 57 & playing trumpet                         \\ \cline{3-4}
                     &                                                         & 58 & playing ukulele                         \\ \cline{3-4}
                     &                                                         & 59 & playing violin                          \\ \cline{3-4}
                     &                                                         & 60 & playing xylophone                       \\ \cline{3-4}
                     &                                                         & 61 & strumming guitar                        \\ \cline{3-4}
                     &                                                         & 62 & tapping guitar                          \\ \hline
\multirow{5}{*}{25}  & \multirow{5}{*}{playing with ball in hand}              & 63 & catching or throwing softball           \\ \cline{3-4}
                     &                                                         & 64 & passing American football (not in game) \\ \cline{3-4}
                     &                                                         & 65 & playing basketball                      \\ \cline{3-4}
                     &                                                         & 66 & playing volleyball                      \\ \cline{3-4}
                     &                                                         & 67 & shooting basketball                     \\ \hline
\multirow{2}{*}{26}  & \multirow{2}{*}{preparing fruit or vegetable}           & 68 & cutting watermelon                      \\ \cline{3-4}
                     &                                                         & 69 & peeling potatoes                        \\ \hline
27                   & punching                                                & 70 & punching person (boxing)                \\ \hline
28                   & pushing                                                 & 71 & pushing wheelchair                      \\ \hline
29                   & reading                                                 & 72 & reading newspaper                       \\ \hline
\multirow{2}{*}{30}  & \multirow{2}{*}{riding animal}                          & 73 & riding mule                             \\ \cline{3-4}
                     &                                                         & 74 & riding or walking with horse            \\ \hline
\multirow{2}{*}{31}  & \multirow{2}{*}{riding on something over water}         & 75 & water sliding                           \\ \cline{3-4}
                     &                                                         & 76 & windsurfing                             \\ \hline
\multirow{5}{*}{32}  & \multirow{5}{*}{riding vehicle}                         & 77 & driving tractor                         \\ \cline{3-4}
                     &                                                         & 78 & riding a bike                           \\ \cline{3-4}
                     &                                                         & 79 & riding mountain bike                    \\ \cline{3-4}
                     &                                                         & 80 & riding unicycle                         \\ \cline{3-4}
                     &                                                         & 81 & using segway                            \\ \hline
33                   & shaving                                                 & 82 & shaving legs                            \\ \hline
\multirow{2}{*}{34}  & \multirow{2}{*}{skiing}                                 & 83 & skiing crosscountry                     \\ \cline{3-4}
                     &                                                         & 84 & skiing slalom                           \\ \hline
35                   & smoking                                                 & 85 & smoking hookah                          \\ \hline
\multirow{2}{*}{36}  & \multirow{2}{*}{swimming}                               & 86 & snorkeling                              \\ \cline{3-4}
                     &                                                         & 87 & swimming butterfly stroke               \\ \hline
37                   & toddler interaction                                     & 88 & crawling baby                           \\ \hline
38                   & tying                                                   & 89 & tying tie                               \\ \hline
\multirow{2}{*}{39}  & \multirow{2}{*}{washing}                                & 90 & washing hair                            \\ \cline{3-4}
                     &                                                         & 91 & washing hands                           \\ \hline
\multirow{2}{*}{40}  & \multirow{2}{*}{waxing}                                 & 92 & waxing eyebrows                         \\ \cline{3-4}
                     &                                                         & 93 & waxing legs                             \\ \hline

\end{longtable}

\subsection{Something-something-v2 - Transferable Activity}

\begin{longtable}{cl}
\caption{List of coarse classes for SSv2-TA}
\label{tab:ssv2_head_class} \\

\hline
\textbf{Coarse ID} & \textbf{Coarse motion} \\
\hline
\endfirsthead

\hline
\textbf{Coarse ID} & \textbf{Coarse motion} \\
\hline
\endhead

\hline
\endfoot

\hline
\endlastfoot

0           & Bending             \\ \hline
1           & Dropping            \\ \hline
2           & Holding             \\ \hline
3           & Letting to roll     \\ \hline
4           & Lifting             \\ \hline
5           & Moving              \\ \hline
6           & Plugging            \\ \hline
7           & Poking              \\ \hline
8           & Pouring             \\ \hline
9           & Pretending          \\ \hline
10          & Pulling             \\ \hline
11          & Pushing             \\ \hline
12          & Putting             \\ \hline
13          & Showing             \\ \hline
14          & Colliding deflected \\ \hline
15          & Falling             \\ \hline
16          & Spilling            \\ \hline
17          & Spinning            \\ \hline
18          & Taking              \\ \hline
19          & Tearing             \\ \hline
20          & Throwing            \\ \hline
21          & Tilting             \\ \hline
22          & Tipping             \\ \hline
23          & Trying but failing  \\ \hline
24          & Turning             \\ \hline
25          & Twisting            \\ \hline
\end{longtable}

\begin{longtable}{clcp{7cm}}

\caption{List of the fine classes in SSv2-TA Set 1}
\label{tab:ssv2_set1_full} \\

\hline
\textbf{Coarse ID} & \textbf{Coarse motion} & \textbf{Fine ID} & \textbf{Coarse motion + fine motion} \\ \hline
\endfirsthead

\hline
\textbf{Coarse ID} & \textbf{Coarse motion} & \textbf{Fine ID} & \textbf{Coarse motion + fine motion} \\ \hline
\endhead

\hline \multicolumn{4}{r}{{Continued on next page}} \\ \hline
\endfoot

\hline
\endlastfoot
\hline
0                    & Bending                              & 0  & Bending something so that it deforms                                             \\ \hline
\multirow{3}{*}{1}   & \multirow{3}{*}{Dropping}            & 1  & Dropping something behind something                                              \\ \cline{3-4}
                     &                                      & 2  & Dropping something in front of something                                         \\ \cline{3-4}
                     &                                      & 3  & Dropping something into something                                                \\ \hline
\multirow{3}{*}{2}   & \multirow{3}{*}{Holding}             & 4  & Holding something                                                                \\ \cline{3-4}
                     &                                      & 5  & Holding something behind something                                               \\ \cline{3-4}
                     &                                      & 6  & Holding something in front of something                                          \\ \hline
\multirow{2}{*}{3}   & \multirow{2}{*}{Letting to roll}     & 7  & Letting something roll along a flat surface                                      \\ \cline{3-4}
                     &                                      & 8  & Letting something roll down a slanted surface                                    \\ \hline
\multirow{4}{*}{4}   & \multirow{4}{*}{Lifting}             & 9  & Lifting a surface with something on it but not enough for it to slide down       \\ \cline{3-4}
                     &                                      & 10 & Lifting a surface with something on it until it starts sliding down              \\ \cline{3-4}
                     &                                      & 11 & Lifting something up completely without letting it drop down                     \\ \cline{3-4}
                     &                                      & 12 & Lifting something up completely, then letting it drop down                       \\ \hline
\multirow{7}{*}{5}   & \multirow{7}{*}{Moving}              & 13 & Moving away from something with your camera                                      \\ \cline{3-4}
                     &                                      & 14 & Moving part of something                                                         \\ \cline{3-4}
                     &                                      & 15 & Moving something across a surface until it falls down                            \\ \cline{3-4}
                     &                                      & 16 & Moving something across a surface without it falling down                        \\ \cline{3-4}
                     &                                      & 17 & Moving something and something away from each other                              \\ \cline{3-4}
                     &                                      & 18 & Moving something and something closer to each other                              \\ \cline{3-4}
                     &                                      & 19 & Moving something and something so they collide with each other                   \\ \hline
6                    & Plugging                             & 20 & Plugging something into something                                                \\ \hline
\multirow{4}{*}{7}   & \multirow{4}{*}{Poking}              & 21 & Poking a hole into some substance                                                \\ \cline{3-4}
                     &                                      & 22 & Poking a hole into something soft                                                \\ \cline{3-4}
                     &                                      & 23 & Poking a stack of something so the stack collapses                               \\ \cline{3-4}
                     &                                      & 24 & Poking a stack of something without the stack collapsing                         \\ \hline
\multirow{2}{*}{8}   & \multirow{2}{*}{Pouring}             & 25 & Pouring something into something                                                 \\ \cline{3-4}
                     &                                      & 26 & Pouring something into something until it overflows                              \\ \hline
\multirow{11}{*}{9}  & \multirow{11}{*}{Pretending}         & 27 & Pretending or failing to wipe something off of something                         \\ \cline{3-4}
                     &                                      & 28 & Pretending or trying and failing to twist something                              \\ \cline{3-4}
                     &                                      & 29 & Pretending to be tearing something that is not tearable                          \\ \cline{3-4}
                     &                                      & 30 & Pretending to close something without actually closing it                        \\ \cline{3-4}
                     &                                      & 31 & Pretending to open something without actually opening it                         \\ \cline{3-4}
                     &                                      & 32 & Pretending to pick something up                                                  \\ \cline{3-4}
                     &                                      & 33 & Pretending to poke something                                                     \\ \cline{3-4}
                     &                                      & 34 & Pretending to pour something out of something, but something is empty            \\ \cline{3-4}
                     &                                      & 35 & Pretending to put something behind something                                     \\ \cline{3-4}
                     &                                      & 36 & Pretending to put something into something                                       \\ \cline{3-4}
                     &                                      & 37 & Pretending to put something next to something                                    \\ \hline
\multirow{4}{*}{10}  & \multirow{4}{*}{Pulling}             & 38 & Pulling something from behind of something                                       \\ \cline{3-4}
                     &                                      & 39 & Pulling something from left to right                                             \\ \cline{3-4}
                     &                                      & 40 & Pulling something from right to left                                             \\ \cline{3-4}
                     &                                      & 41 & Pulling something onto something                                                 \\ \hline
\multirow{5}{*}{11}  & \multirow{5}{*}{Pushing}             & 42 & Pushing something from left to right                                             \\ \cline{3-4}
                     &                                      & 43 & Pushing something from right to left                                             \\ \cline{3-4}
                     &                                      & 44 & Pushing something off of something                                               \\ \cline{3-4}
                     &                                      & 45 & Pushing something onto something                                                 \\ \cline{3-4}
                     &                                      & 46 & Pushing something so it spins                                                    \\ \hline
\multirow{10}{*}{12} & \multirow{10}{*}{Putting}            & 47 & Putting number of something onto something                                       \\ \cline{3-4}
                     &                                      & 48 & Putting something and something on the table                                     \\ \cline{3-4}
                     &                                      & 49 & Putting something behind something                                               \\ \cline{3-4}
                     &                                      & 50 & Putting something in front of something                                          \\ \cline{3-4}
                     &                                      & 51 & Putting something into something                                                 \\ \cline{3-4}
                     &                                      & 52 & Putting something next to something                                              \\ \cline{3-4}
                     &                                      & 53 & Putting something on a flat surface without letting it roll                      \\ \cline{3-4}
                     &                                      & 54 & Putting something on a surface                                                   \\ \cline{3-4}
                     &                                      & 55 & Putting something on the edge of something so it is not supported and falls down \\ \cline{3-4}
                     &                                      & 56 & Putting something onto a slanted surface but it doesn't glide down               \\ \hline
\multirow[t]{4}{*}{13}  & \multirow[t]{4}{*}{Showing}             & 57 & Showing a photo of something to the camera                                       \\ \cline{3-4}
                     &                                      & 58 & Showing something behind something                                               \\ \cline{3-4}
                     &                                      & 59 & Showing something next to something                                              \\ \cline{3-4}
                     &                                      & 60 & Showing something on top of something                                            \\ \hline
\multirow{2}{*}{14}  & \multirow{2}{*}{Colliding deflected} & 61 & Something being deflected from something                                         \\ \cline{3-4}
                     &                                      & 62 & Something colliding with something and both are being deflected                  \\ \hline
15                   & Falling                              & 63 & Something falling like a feather or paper                                        \\ \hline
\multirow{2}{*}{16}  & \multirow{2}{*}{Spilling}            & 64 & Spilling something behind something                                              \\ \cline{3-4}
                     &                                      & 65 & Spilling something next to something                                             \\ \hline
17                   & Spinning                             & 66 & Spinning something so it continues spinning                                      \\ \hline
\multirow{2}{*}{18}  & \multirow{2}{*}{Taking}              & 67 & Taking one of many similar things on the table                                   \\ \cline{3-4}
                     &                                      & 68 & Taking something from somewhere                                                  \\ \hline
19                   & Tearing                              & 69 & Tearing something into two pieces                                                \\ \hline
\multirow{3}{*}{20}  & \multirow{3}{*}{Throwing}            & 70 & Throwing something                                                               \\ \cline{3-4}
                     &                                      & 71 & Throwing something against something                                             \\ \cline{3-4}
                     &                                      & 72 & Throwing something in the air and catching it                                    \\ \hline
21                   & Tilting                              & 73 & Tilting something with something on it slightly so it doesn't fall down          \\ \hline
22                   & Tipping                              & 74 & Tipping something over                                                           \\ \hline
\multirow{2}{*}{23}  & \multirow{2}{*}{Trying but failing}  & 75 & Trying but failing to attach something to something because it doesn't stick     \\ \cline{3-4}
                     &                                      & 76 & Trying to bend something unbendable so nothing happens                           \\ \hline
\multirow{3}{*}{24}  & \multirow{3}{*}{Turning}             & 77 & Turning something upside down                                                    \\ \cline{3-4}
                     &                                      & 78 & Turning the camera downwards while filming something                             \\ \cline{3-4}
                     &                                      & 79 & Turning the camera left while filming something                                  \\ \hline
25                   & Twisting                             & 80 & Twisting (wringing) something wet until water comes out                          \\ \hline
\end{longtable}

\begin{longtable}{clcp{7cm}}

\caption{List of the fine classes in SSv2-TA Set 2}
\label{tab:ssv2_set2_full} \\

\hline
\textbf{Coarse ID} & \textbf{Coarse motion} & \textbf{Fine ID} & \textbf{Coarse motion + fine motion} \\ \hline
\endfirsthead

\hline
\textbf{Coarse ID} & \textbf{Coarse motion} & \textbf{Fine ID} & \textbf{Coarse motion + fine motion} \\ \hline
\endhead

\hline \multicolumn{4}{r}{{Continued on next page}} \\ \hline
\endfoot

\hline
\endlastfoot
\hline
0                   & Bending                      & 0  & Bending something until it breaks                                                                  \\ \hline
\multirow{2}{*}{1}  & \multirow{2}{*}{Dropping}    & 1  & Dropping something next to something                                                               \\ \cline{3-4}
                    &                              & 2  & Dropping something onto something                                                                  \\ \hline
\multirow{2}{*}{2}  & \multirow{2}{*}{Holding}     & 3  & Holding something next to something                                                                \\ \cline{3-4}
                    &                              & 4  & Holding something over something                                                                   \\ \hline
3                   & Letting to roll              & 5  & Letting something roll up a slanted surface, so it rolls back down                                 \\ \hline
\multirow{3}{*}{4}  & \multirow{3}{*}{Lifting}     & 6  & Lifting something with something on it                                                             \\ \cline{3-4}
                    &                              & 7  & Lifting up one end of something without letting it drop down                                       \\ \cline{3-4}
                    &                              & 8  & Lifting up one end of something, then letting it drop down                                         \\ \hline
\multirow{7}{*}{5}  & \multirow{7}{*}{Moving}      & 9  & Moving something and something so they pass each other                                             \\ \cline{3-4}
                    &                              & 10 & Moving something away from something                                                               \\ \cline{3-4}
                    &                              & 11 & Moving something away from the camera                                                              \\ \cline{3-4}
                    &                              & 12 & Moving something closer to something                                                               \\ \cline{3-4}
                    &                              & 13 & Moving something down                                                                              \\ \cline{3-4}
                    &                              & 14 & Moving something towards the camera                                                                \\ \cline{3-4}
                    &                              & 15 & Moving something up                                                                                \\ \hline
6                   & Plugging                     & 16 & Plugging something into something but pulling it right out as you remove your hand                 \\ \hline
\multirow{4}{*}{7}  & \multirow{4}{*}{Poking}      & 17 & Poking something so it slightly moves                                                              \\ \cline{3-4}
                    &                              & 18 & Poking something so lightly that it doesn't or almost doesn't move                                 \\ \cline{3-4}
                    &                              & 19 & Poking something so that it falls over                                                             \\ \cline{3-4}
                    &                              & 20 & Poking something so that it spins around                                                           \\ \hline
\multirow{2}{*}{8}  & \multirow{2}{*}{Pouring}     & 21 & Pouring something onto something                                                                   \\ \cline{3-4}
                    &                              & 22 & Pouring something out of something                                                                 \\ \hline
\multirow{11}{*}{9} & \multirow{11}{*}{Pretending} & 23 & Pretending to put something on a surface                                                           \\ \cline{3-4}
                    &                              & 24 & Pretending to put something onto something                                                         \\ \cline{3-4}
                    &                              & 25 & Pretending to put something underneath something                                                   \\ \cline{3-4}
                    &                              & 26 & Pretending to scoop something up with something                                                    \\ \cline{3-4}
                    &                              & 27 & Pretending to spread air onto something                                                            \\ \cline{3-4}
                    &                              & 28 & Pretending to sprinkle air onto something                                                          \\ \cline{3-4}
                    &                              & 29 & Pretending to squeeze something                                                                    \\ \cline{3-4}
                    &                              & 30 & Pretending to take something from somewhere                                                        \\ \cline{3-4}
                    &                              & 31 & Pretending to take something out of something                                                      \\ \cline{3-4}
                    &                              & 32 & Pretending to throw something                                                                      \\ \cline{3-4}
                    &                              & 33 & Pretending to turn something upside down                                                           \\ \hline
\multirow{4}{*}{10} & \multirow{4}{*}{Pulling}     & 34 & Pulling something out of something                                                                 \\ \cline{3-4}
                    &                              & 35 & Pulling two ends of something but nothing happens                                                  \\ \cline{3-4}
                    &                              & 36 & Pulling two ends of something so that it gets stretched                                            \\ \cline{3-4}
                    &                              & 37 & Pulling two ends of something so that it separates into two pieces                                 \\ \hline
\multirow{4}{*}{11} & \multirow{4}{*}{Pushing}     & 38 & Pushing something so that it almost falls off but doesn't                                          \\ \cline{3-4}
                    &                              & 39 & Pushing something so that it falls off the table                                                   \\ \cline{3-4}
                    &                              & 40 & Pushing something so that it slightly moves                                                        \\ \cline{3-4}
                    &                              & 41 & Pushing something with something                                                                   \\ \hline
\multirow{9}{*}{12} & \multirow{9}{*}{Putting}     & 42 & Putting something onto something                                                                   \\ \cline{3-4}
                    &                              & 43 & Putting something onto something else that cannot support it so it falls down                      \\ \cline{3-4}
                    &                              & 44 & Putting something similar to other things that are already on the table                            \\ \cline{3-4}
                    &                              & 45 & Putting something that can't roll onto a slanted surface, so it slides down                        \\ \cline{3-4}
                    &                              & 46 & Putting something that can't roll onto a slanted surface, so it stays where it is                  \\ \cline{3-4}
                    &                              & 47 & Putting something that cannot actually stand upright upright on the table, so it falls on its side \\ \cline{3-4}
                    &                              & 48 & Putting something underneath something                                                             \\ \cline{3-4}
                    &                              & 49 & Putting something upright on the table                                                             \\ \cline{3-4}
                    &                              & 50 & Putting something, something and something on the table                                            \\ \hline
\multirow[t]{3}{*}{13} & \multirow[t]{3}{*}{Showing}     & 51 & Showing something to the camera                                                                    \\ \cline{3-4}
                    &                              & 52 & Showing that something is empty                                                                    \\ \cline{3-4}
                    &                              & 53 & Showing that something is inside something                                                         \\ \hline
14                  & Colliding deflected          & 54 & Something colliding with something and both come to a halt                                         \\ \hline
15                  & Falling                      & 55 & Something falling like a rock                                                                      \\ \hline
16                  & Spilling                     & 56 & Spilling something onto something                                                                  \\ \hline
17                  & Spinning                     & 57 & Spinning something that quickly stops spinning                                                     \\ \hline
18                  & Taking                       & 58 & Taking something out of something                                                                  \\ \hline
19                  & Tearing                      & 59 & Tearing something just a little bit                                                                \\ \hline
\multirow{2}{*}{20} & \multirow{2}{*}{Throwing}    & 60 & Throwing something in the air and letting it fall                                                  \\ \cline{3-4}
                    &                              & 61 & Throwing something onto a surface                                                                  \\ \hline
21                  & Tilting                      & 62 & Tilting something with something on it until it falls off                                          \\ \hline
22                  & Tipping                      & 63 & Tipping something with something in it over, so something in it falls out                          \\ \hline
23                  & Trying but failing           & 64 & Trying to pour something into something, but missing so it spills next to it                       \\ \hline
\multirow{2}{*}{24} & \multirow{2}{*}{Turning}     & 65 & Turning the camera right while filming something                                                   \\ \cline{3-4}
                    &                              & 66 & Turning the camera upwards while filming something                                                 \\ \hline
25                  & Twisting                     & 67 & Twisting something                                                                                 \\ \hline
\end{longtable}

\twocolumn

\end{document}